\documentclass{article}

\usepackage{arxiv}
\usepackage[framemethod=tikz]{mdframed}
\usepackage{lipsum}
\usepackage{hyperref}
\usepackage{graphicx}
\usepackage{float}
\usepackage{multicol}
\usepackage[ruled]{algorithm2e}
\usepackage{comment}

\usepackage{xcolor}
\usepackage{soul}
\usepackage[caption=false]{subfig}
\usepackage{multirow}
\usepackage{mathcomp}
\usepackage{cite}
\title{Incremental Learning in Deep Convolutional Neural Networks Using Partial Network Sharing}

\author{
 Syed Shakib Sarwar \\
  Department of Electrical and Computer Engineering\\
  Purdue University\\
  West Lafayette, IN, 47906 \\
  \texttt{sarwar@purdue.edu} \\
   \And
  Aayush Ankit\\
  Department of Electrical and Computer Engineering\\
  Purdue University\\
  West Lafayette, IN, 47906 \\
  \texttt{aankit@purdue.edu} \\
   \And
 Kaushik Roy \\
  Department of Electrical and Computer Engineering\\
  Purdue University\\
  West Lafayette, IN, 47906 \\
  \texttt{kaushik@purdue.edu} \\
}

\begin{document}
\maketitle
%

%
%
%



\begin{abstract}
Deep convolutional neural network (DCNN) based supervised learning is a widely practiced approach for large-scale image classification.  However, retraining these large networks to accommodate new, previously unseen data demands high computational time and energy requirements. Also, previously seen training samples may not be available at the time of retraining. We propose an efficient training methodology and incrementally growing DCNN to learn new tasks while sharing part of the base network. Our proposed methodology is inspired by transfer learning techniques, although it does not forget previously learned tasks. An updated network for learning new set of classes is formed using previously learned convolutional layers (shared from initial part of base network) with addition of few newly added convolutional kernels included in the later layers of the network. We employed a `clone-and-branch' technique which allows the network to learn new tasks one after another without any performance loss in old tasks. We evaluated the proposed scheme on several recognition applications. The classification accuracy achieved by our approach is comparable to the regular incremental learning approach (where networks are updated with new training samples only, without any network sharing), while achieving energy efficiency, reduction in storage requirements, memory access and training time.
\end{abstract}



\section{Introduction}

Deep Convolutional Neural Networks (DCNNs) have achieved remarkable success in various cognitive applications, particularly in computer vision \cite{1}. They have shown human like performance on a variety of recognition, classification and inference tasks, albeit at a much higher energy consumption. One of the major challenges for convolutional networks is the computational complexity and the time needed to train large networks. Since training of DCNNs requires state-of-the-art accelerators like GPUs ~\cite{chetlur2014cudnn}, large training overhead has restricted the usage of DCNNs to clouds and servers. It is common to pre-train a DCNN on a large dataset (e.g. ImageNet, which contains 1.2 million images with 1000 categories), and then use the trained network either as an initialization or a fixed feature extractor for the specific application  ~\cite{sharif2014cnn}. A major downside of such DCNNs is the inability to learn new information since the learning process is static and only done once before it is exposed to practical applications. In real-world scenarios, classes and their associated labeled data are always collected in an incremental manner. To ensure applicability of DCNNs in such cases, the learning process needs to be continuous. However, retraining these large networks using both previously seen and unseen data to accommodate new data, is not feasible most of the time. The training samples for already learned classes may be proprietary, or simply too cumbersome to use in training a new task. Also, to ensure data privacy, training samples should be discarded after use. Incremental learning plays a critical role in alleviating these issues by ensuring continuity in the learning process through regular model update based only on the new available batch of data. Nevertheless, incremental learning can be computationally expensive and time consuming, if the network is large enough.
 
\par
This paper focuses on incremental learning on deep convolutional neural network (DCNN) for image classification task. In doing so, we attempt to address the more fundamental issue: an efficient learning system must deal with new knowledge that it is exposed to, as humans do. To achieve this goal, there are two major challenges. First, as new data becomes available, we should not start learning from scratch. Rather, we leverage what we have already learned and combine them with new knowledge in a continuous manner. Second, to accommodate new data, if there is a need to increase the capacity of our network, we will have to do it in an efficient way. 
We would like to clarify that incremental learning is not a replacement of regular training. In the regular case, samples for all classes are available from the beginning of training. However, in incremental learning, sample data corresponding to new tasks become available after the base network is already trained and sample data for already learned task are no longer available for retraining the network to learn all tasks (old and new) simultaneously.
Our approach to incremental learning is similar to transfer learning ~\cite{pan2010survey} and domain adaptation methods ~\cite{patricia2014learning}. Transfer learning utilizes knowledge acquired from one task assisting to learn another. Domain adaptation transfers the knowledge acquired for a task from a dataset to another (related) dataset. These paradigms are very popular in computer vision. Though incremental learning is similar in spirit to transfer, multi-task, and lifelong learning; so far, no work has provided a perfect solution to the problem of continuously adding new tasks based on adapting shared parameters without access to training data for previously learned tasks.\par

There have been several prior works on incremental learning of neural networks. Many of them focus on learning new tasks from fewer samples \cite{fei2006one,lampert2009learning} utilizing transfer learning techniques. To avoid learning new categories from scratch, Fei-Fei et al. \cite{fei2006one} proposed a Bayesian transfer learning method using very few training samples. By introducing attribute-based classification the authors \cite{lampert2009learning} achieved zero-shot learning (learning a new class from zero labeled samples). These works rely on shallow models instead of DCNN, and the category size is small in comparison. The challenge of applying incremental learning (transfer learning as well) on DCNN lies in the fact that it consists of both feature extractor and classifier in one architecture. Polikar et al. \cite{polikar2001learn++} utilized ensemble of classifiers by generating multiple hypotheses using training data sampled according to carefully tailored distributions. The outputs of the resulting classifiers are combined using a weighted majority voting procedure. This method can handle an increasing number of classes, but needs training data for all classes to occur repeatedly. Inspired form Polikar et al. \cite{polikar2001learn++}, Medera and Babinec ~\cite{medera2009incremental} utilized ensemble of modified convolutional neural networks as classifiers by generating multiple hypotheses. The existing classifiers are improved in ~\cite{polikar2001learn++, medera2009incremental} by combining new hypothesis generated from newly available examples without compromising classification performance on old data. The new data in ~\cite{polikar2001learn++, medera2009incremental} may or may not contain new classes.
Another method by Royer and Lampert ~\cite{royer2015classifier} can adapt classifiers to a time-varying data stream. However, the method is unable to handle new classes. Pentina et al. ~\cite{pentina2015curriculum} have shown that learning multiple tasks sequentially can improve classification accuracy. Unfortunately, for choosing the sequence, the data for all tasks must be available to begin with.
Xiao et al. ~\cite{xiao2014error} proposed a training algorithm that grows a network not only incrementally but also hierarchically. In this tree-structured model, classes are grouped according to similarities, and self-organized into different levels of the hierarchy. All new networks are cloned from existing ones and therefore inherit learned features. These new networks are fully retrained and connected to base network. The problem with this method is the increase of hierarchical levels as new set of classes are added over time. Another hierarchical approach was proposed in Roy et al. \cite{roy2018tree} where the network grows in a tree-like manner to accommodate the new classes. However, in this approach, the root node of the tree structure is retrained with all training samples (old and new classes) during growing the network.\par 
 
Li and Hoiem ~\cite{li2016learning} proposed `Learning without Forgetting' (LwF) to incrementally train a single network to learn multiple tasks. Using only examples for the new task, the authors optimize both for high accuracy for the new task and for preservation of responses on the existing tasks from the original network. Though only the new examples were used for training, the whole network must be retrained every time a new task needs to be learned. Recently, Rebuffi et al. ~\cite{rebuffi2016icarl} addressed some of the drawbacks in  ~\cite{li2016learning} with their decoupled classifier and representation learning approach. However, they rely on a subset of the original training data to preserve the performance on the old classes. Shmelkov et al. ~\cite{shmelkov2017incremental} proposed a solution by forming a loss function to balance the interplay between predictions on the new classes and a new distillation loss which minimizes the discrepancy between responses for old classes from the original and the updated networks. This method can be performed multiple times, for a new set of classes in each step. However, every time it incurs a moderate drop in performance compared to the baseline network trained on the ensemble of data. Also, the whole process has substantial overhead in terms of compute energy and memory.\par
Another way to accommodate new classes is growing the capacity of the network with new layers ~\cite{rusu2016progressive}, selectively applying strong per-parameter regularization ~\cite{kirkpatrick2017overcoming}. The drawbacks to these methods are the rapid increase in the number of new parameters to be learned ~\cite{rusu2016progressive}, and they are more suited to reinforcement learning ~\cite{kirkpatrick2017overcoming}. Aljundi et al. ~\cite{aljundi2016expert} proposed a gating approach to select the model that can provide the best performance for the current task. It introduces a set of gating auto-encoders that learn a representation for the task at hand, and, at test time, automatically forward the test sample to the relevant expert. This method performs very well on image classification and video prediction problems. However, the training of autoencoders for each task requires significant effort.
Incremental learning is also explored in Spiking Neural Networks (SNN) domain. An unsupervised learning mechanism is proposed by Panda et al. \cite{panda2018asp} for improved recognition with SNNs for on-line learning in a dynamic environment. This mechanism helps in gradual forgetting of insignificant data while retaining significant, yet old, information thus trying to addresses catastrophic forgetting.\par
In the context of incremental learning, most work has focused on how to exploit knowledge from previous tasks and transfer it to a new task. Little attention has gone to the related and equally important problem of hardware and energy requirements for model update. Our work differs in goal, as we want to grow a DCNN with reduced effort to accommodate new tasks (sets of classes) by network sharing, without forgetting the old tasks (sets of classes). The key idea of this work is the unique `clone-and-branch' technique which allows the network to learn new tasks one after another without any performance loss in old tasks. Cloning layers provides a good starting point for learning a new task compared to randomly initialized weights. The kernels learn quickly, and training converges faster. It allows us to employ fine-tuning in the new branch, saving training energy and time compared to training from scratch. On the other hand, branching allows the network to remember task specific weight parameters, hence the network does not forget old tasks (in task specific scenario) no matter how many new tasks it has learned. The novelty of this work lies in the fact that we developed an empirical mechanism to identify how much of the network can be shared as new tasks are learned. We also quantified the energy consumption, training time and memory storage savings associated with models trained with different amounts of sharing to emphasize the importance of network sharing from hardware point of view. 
Our proposed method is unique since it does not require any algorithmic changes and can be implemented in any existing hardware if additional memory is available for the supplementary parameters needed to learn new classes. There is no overhead of storing any data sample or statistical information of the learned classes. It also allows on-chip model update using a programmable instruction cache. Many of the state-of-the-art DNN accelerators support this feature. However, FPGAs are the kind of hardware architecture that is best suited for the proposed method. It offers highly flexible micro-architecture with reusable functional modules and additional memory blocks in order to account for dynamic changes.\par 
In summary, the key contributions of our work are as follows:

\begin{figure*}[h]
  \centering
  \includegraphics[width=0.6\textwidth]{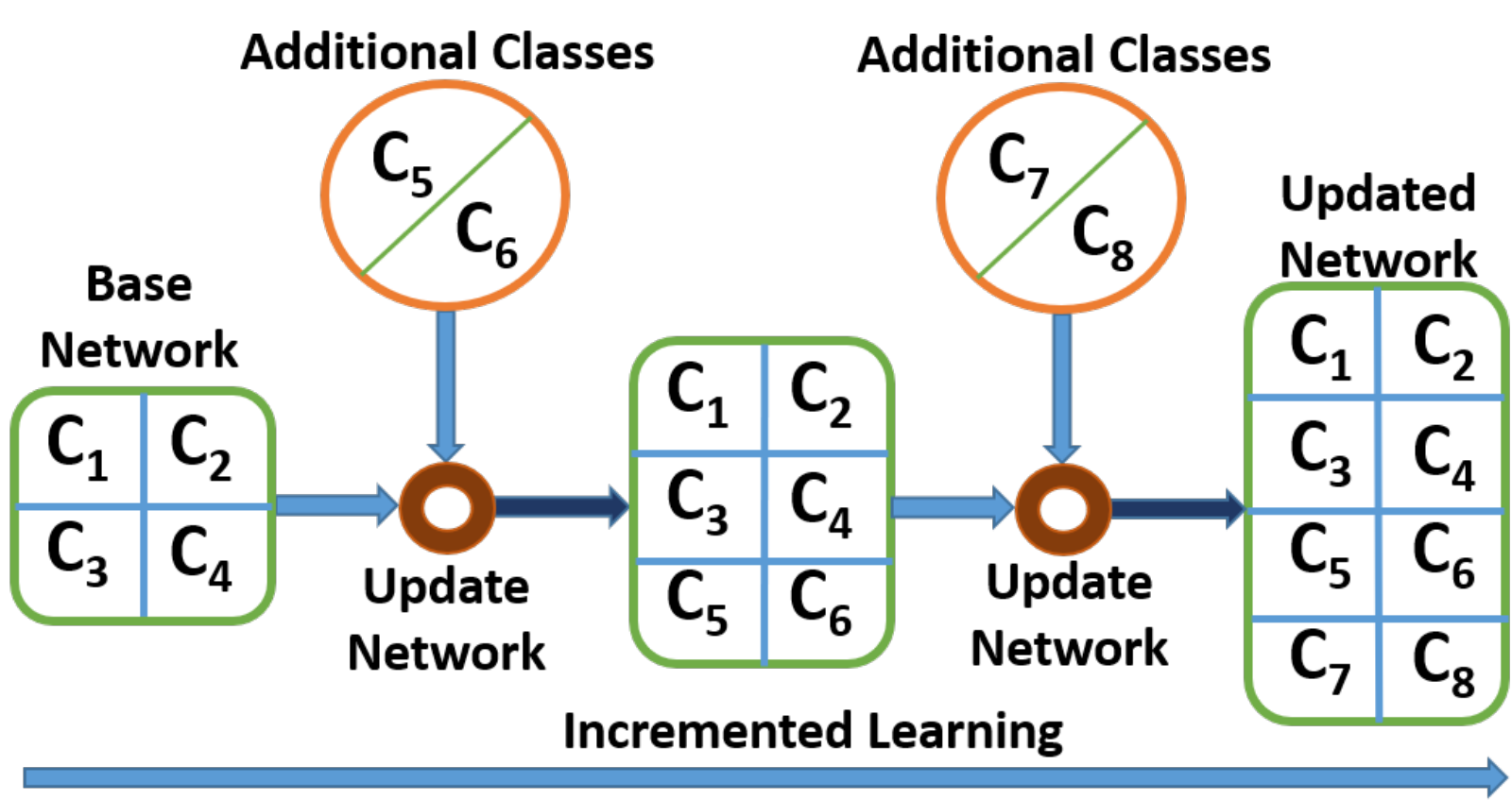}
  \caption{Incremental learning model: the network needs to grow its capacity with arrival of data of new classes.}
  \label{fig1}
\end{figure*} 

\begin{itemize}
  \item	We propose sharing of convolutional layers to reduce computational complexity while training a network to accommodate new tasks (sets of classes) without forgetting old tasks (sets of classes).
  \item We developed a methodology to identify optimal sharing of convolutional layers in order to get the best trade-off between accuracy and other parameters of interest, especially energy consumption, training time and memory access.
  \item We developed a cost estimation model for quantifying energy consumption of the network during training, based on the \underline{M}ultiplication and \underline{Ac}cumulation (MAC) operations and number of memory access in the training algorithm.
  \item We substantiate the scalability and robustness of the proposed methodology by applying the proposed method to different network structures trained for different benchmark datasets. 
\end{itemize}
We show that our proposed methodology leads to energy efficiency, reduction in storage requirements, memory access and training time, while maintaining classification accuracy without accessing training samples of old tasks.

\section{Incremental Learning}
A crude definition of incremental learning is that it is a continuous learning process as batches of labeled data of new classes are gradually made available. In literature, the term ``incremental learning'' is also referred to incremental network growing and pruning or on-line learning. Moreover, various other terms, such as lifelong learning, constructive learning and evolutionary learning have also been used to denote learning new information. Development of a pure incremental learning model is important in mimicking real, biological brains. Owing to superiority of biological brain, humans and other animals can learn new events without forgetting old events. However, exact sequential learning does not work flawlessly in artificial neural networks. The reasons can be the use of a fixed architecture and/or a training algorithm based on minimizing an objective function which results in {\it ``catastrophic interference''}. It is due to the fact that the minima of the objective function for one example set may be different from the minima for subsequent example sets. Hence each successive training set causes the network to partially or completely forget previous training sets. This problem is called the {\it``stability-plasticity dilemma''} ~\cite{mermillod2013stability}. To address these issues, we define an incremental learning algorithm that meets the following criteria:\\
\textbf{i.} It should be able to grow the network and accommodate new tasks (sets of classes) that are introduced with new examples.\newline
\textbf{ii.} Training for new tasks (sets of classes) should have minimal overhead.\newline
\textbf{iii.} It should not require access to the previously seen examples used to train the existing classifier.\newline
\textbf{iv.} It should preserve previously acquired knowledge,  {\it i.e.} it should not suffer from catastrophic forgetting.\par

In this work, we developed an efficient training methodology that can cover aforementioned criteria. Let us comprehend the concept with a simple example. 
Assume that a base network is trained with four classes of task 0 ($C_{1}-C_{4}$), and all training data of those four classes are discarded after training. Next, sample data for task 1 with two classes (($C_{5},C_{6}$)) arrive and the network needs to accommodate them while keeping knowledge of the initial four classes. Hence, the network capacity has to be increased and the network has to be retrained with only the new data of task 1 (of $C_{5}$ and $C_{6}$) in an efficient way so that the updated network can classify both task's classes ($C_{1}-C_{6}$).
If the tasks are classified separately, then it is a task specific classification. On the other hand, when they are classified together, then it is called combined classification. We will primarily focus on the task specific scenario while also considering the combined classification. Figure \ref{fig1} shows the overview of the incremental learning model we use.

\subsection{Advantages}
There are several major benefits of incremental learning. 
\subsubsection{Enable training in low power devices} 
Training a deep network from scratch requires enormous amount of time and energy which is not affordable for low power devices (embeded systems, mobile devices, IoTs etc.). Therefore, a deep network is trained off-chip and deployed in the edge devices. When data for new task are available, it can not be used for learning in the device because of two reasons; i) the device does not have access to sample data for already learned tasks, ii) the device does not posses the capability to retrain the whole network. However, the new tasks can be learned incrementally by reusing knowledge from existing network without requiring data samples of old tasks. This enables the low power devices to update the existing network by incrementally retraining it within their power budget and hardware limitations. 
\subsubsection{Speed up model update}
If knowledge from existing network can be reused while learning new tasks (with new data samples only) without forgetting old tasks, then the updating process of an existing network will be very fast.
\subsubsection{Ensure data privacy}
Incremental learning do not require access to old training data. Therefore, all training samples can be discarded after each training session, which will disallow misuse of private data.    
\subsubsection{Reduce storage requirements}
Deep networks require humongous amount of data to train. Since training data samples are not required to be stored for incremental learning, the storage requirement for updating a network is greatly reduced. 

The following section will describe the design approach of the proposed scheme.

\section{Design Approach}
The superiority of DCNNs comes from the fact that it contains both feature extractor and classifier in the same network with many layers. `\textit{Sharing}' convolutional layers as fixed feature extractors is the base of our proposed training methodology. `\textit{Sharing}' means reusing already learned network parameters/layers to learn new set of classes. Note that in all cases, while learning new classes, only newly available data is used.  
Also, we assume that new classes will have similar features as the old classes. Therefore, we separate a single dataset into several sets so that they can be used as old and new data while updating the network. All accuracies reported in this work are test accuracies (training samples and test samples are mutually exclusive).\par 
This section outlines the key ideas behind the proposed methodology.

\begin{figure}[ht]
  \centering
  \includegraphics[width=0.75\textwidth]{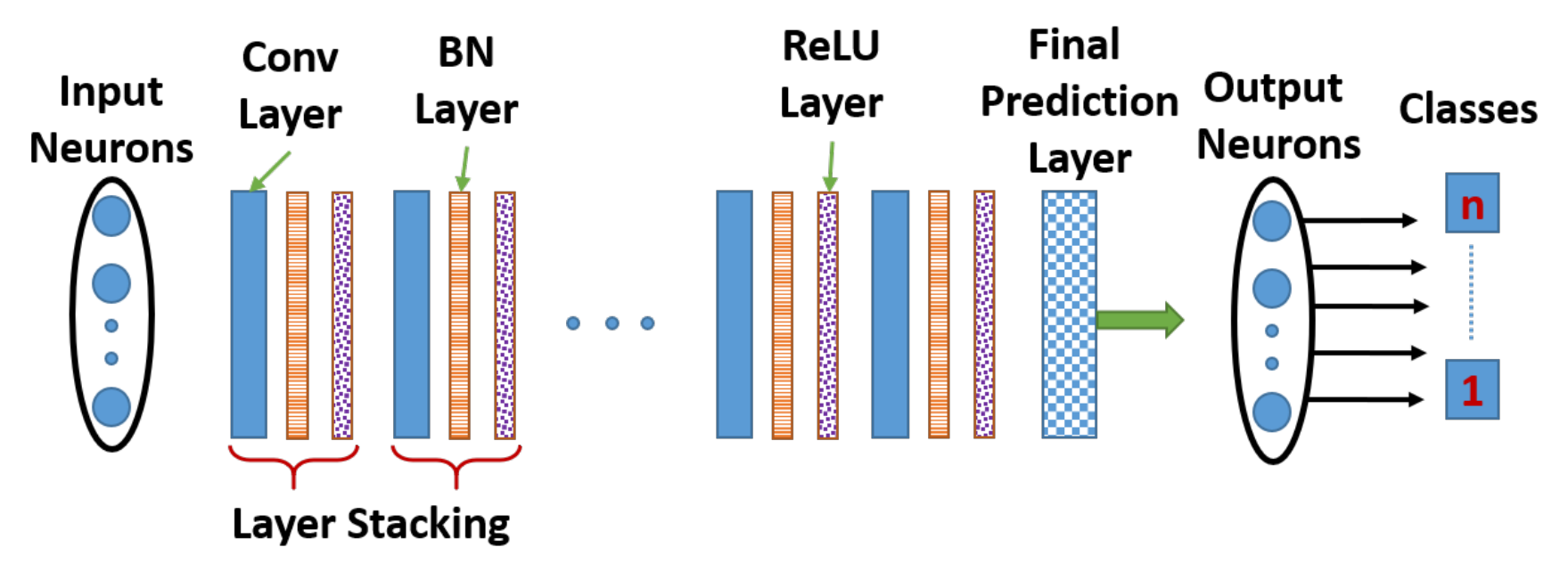}
  \caption{The ResNet~\cite{he2016deep} network structure used for implementing large scale DCNN. For simplicity, the skip connections of ResNet architecture is not shown here.}
  \label{fig36}
\end{figure} 

\begin{figure*}[h]
\centering
\subfloat[]{\label{fig37a}
  \centering
  \includegraphics[width=.5\textwidth]{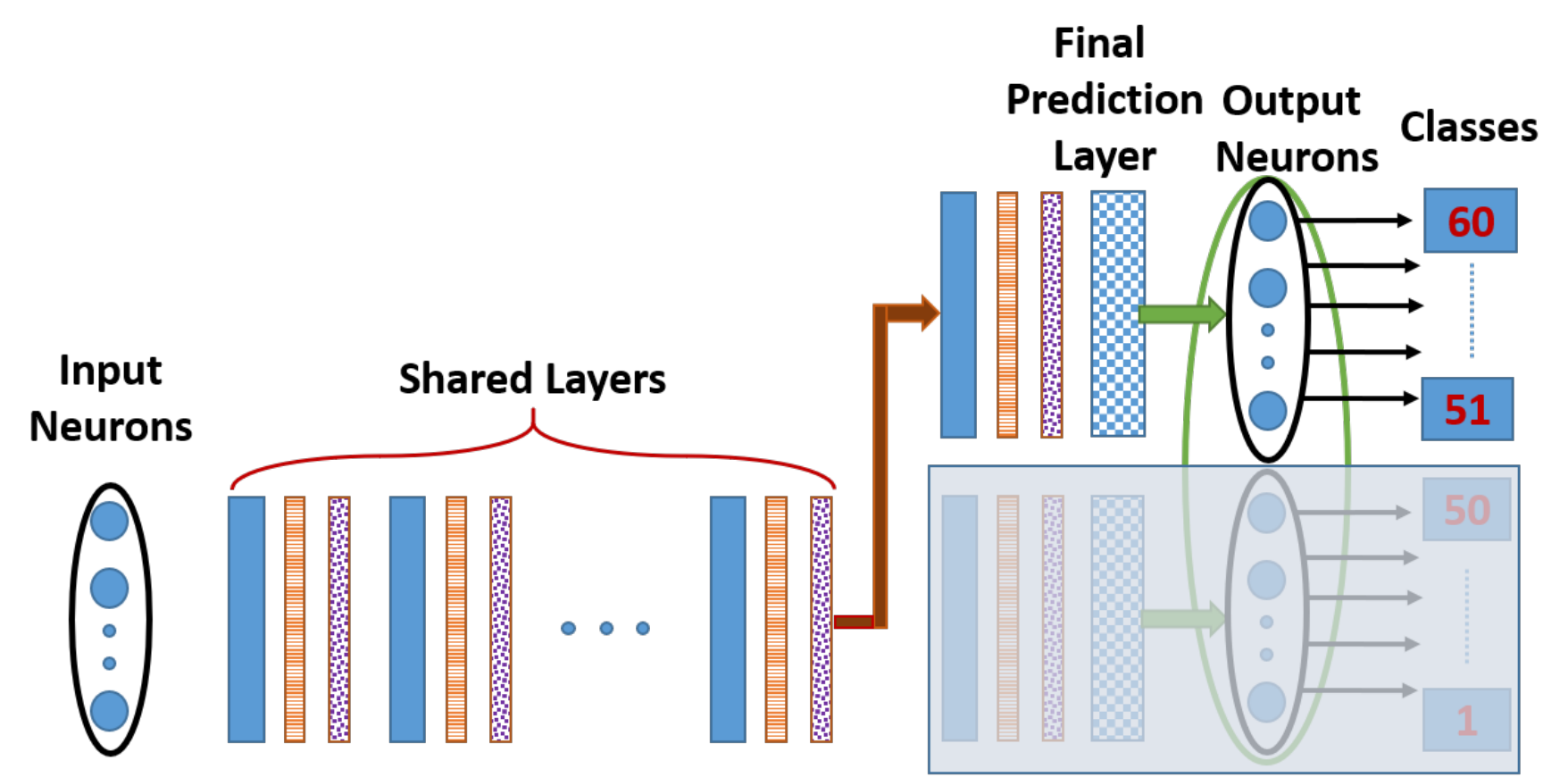}}
\subfloat[]{\label{fig37b}
  \centering
  \includegraphics[width=.5\textwidth]{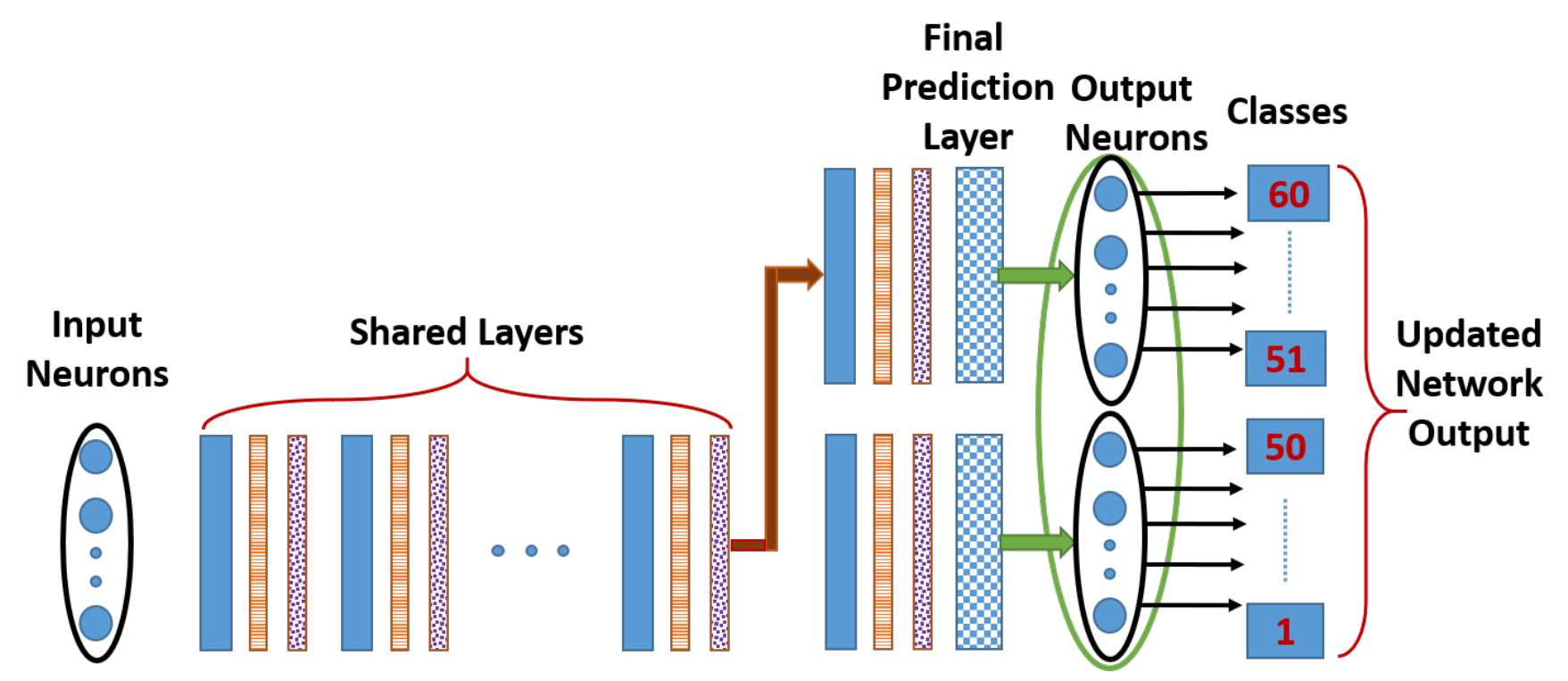}}
\caption{(a) Incremental training for accommodating new classes in the base network. The parameters of the shared layers are frozen. The semi-transparent rectangle implies that the part is disconnected from training. The new convolutional layer is cloned from the base network and only that part is retrained with the new data samples for the new classes, while the last convolutional layer of the base network remain disconnected.  
(b) After retraining the cloned layer, we add it to the existing network as a new branch, and form the updated network.}
\label{fig37}
\end{figure*}

\subsection{Replacing Part of the Base Network with New Convolutional Layers}
A large DCNN usually has many convolutional layers followed by a fully connected final classifier. To apply our approach in a large DCNN, we implemented ResNet ~\cite{he2016deep} for a real-world object recognition application. The network structure is depicted in Figure \ref{fig36}. CIFAR-100 ~\cite{krizhevsky2009learning} was used as the benchmark dataset. We trained a base network (ResNet50), with 50 classes out of the 100 in CIFAR-100. Then we added rest of the 50 classes to the existing network in three installments of 20, 20 and 10. The classes for forming the sets were chosen randomly and each set is mutually exclusive. Each time when we update the network for new tasks, we clone the last convolutional layer and following classifier layers, while sharing the initial convolutional layers from the base network, and retrain it using examples for the new set of classes only (Figure \ref{fig37a}). After retraining the cloned branch, we add it to the existing network as a new branch as shown in Figure \ref{fig37b}. Note that, the initial part of the base network is shared and frozen. After training the branch network for new task (additional classes), we have the updated network that can do task specific classification as well as combined classification.
During task specific classification, only the task specific branch will be active, while for combined classification all branches will be active at the same time. Since during training the new branches, shared and old task specific parameters are not altered, the network will not forget already learned tasks. However, training only 1 conv layer and classifier layers is not enough to learn a new task properly. Hence, new task performance suffers in this configuration. 
We compared the accuracies achieved by this method with the accuracy of a network of same depth, trained without sharing any learning parameter, and observed that there is an 8-12\% accuracy degradation for the former method. We also assessed the updated network accuracy for the all 100 classes by generating prediction probabilities from each of the separately trained networks and selecting the maximum probability. Even for the updated network, we observed about 10\% accuracy degradation. To counter this accuracy degradation, we developed a training methodology that will be described in the following subsection.

\begin{figure*}[h]
\centering
\subfloat[]{\label{fig38a}
  \centering
  \includegraphics[width=.5\textwidth]{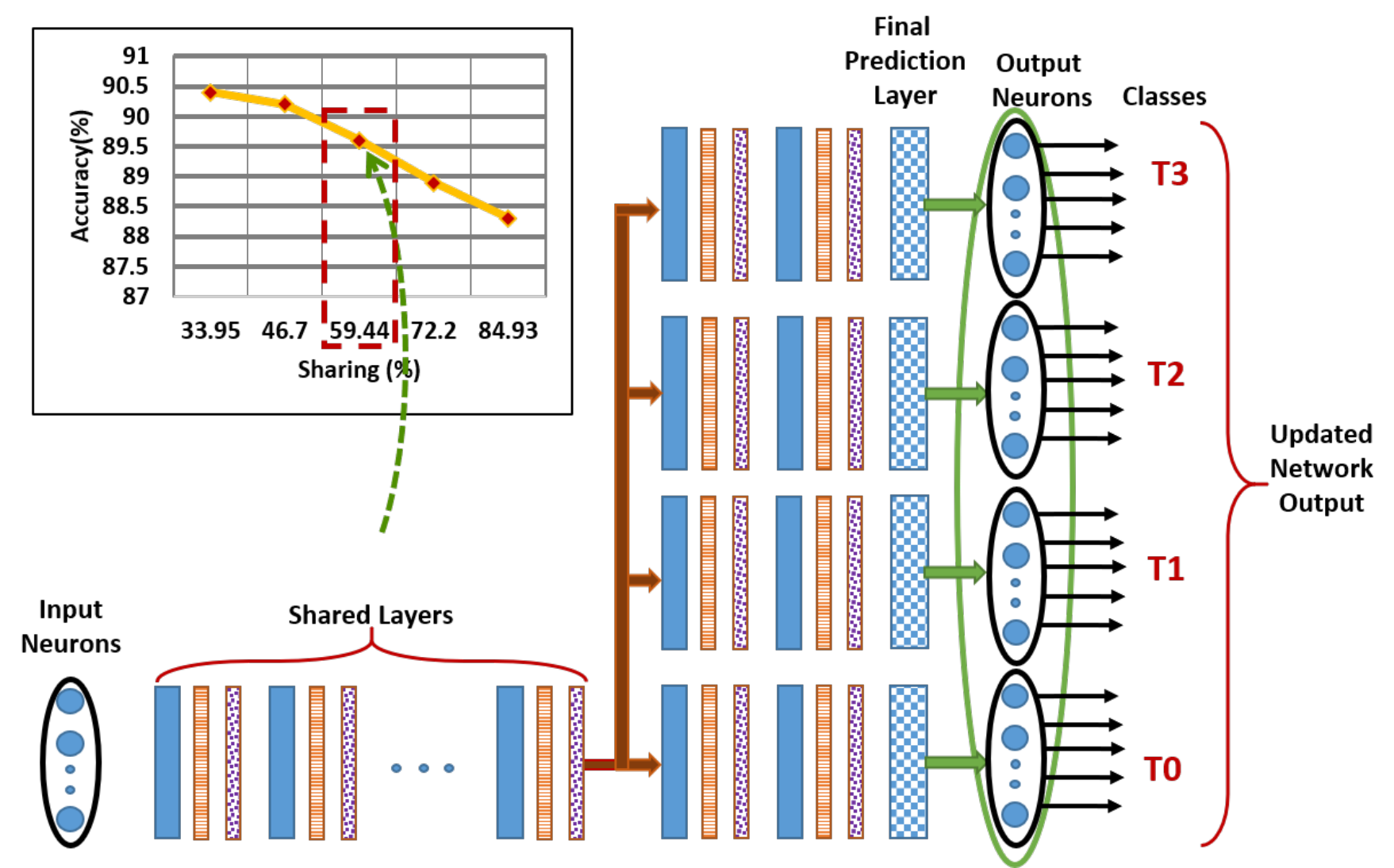}}
\subfloat[]{\label{fig38b}
  \centering
  \includegraphics[width=.5\textwidth]{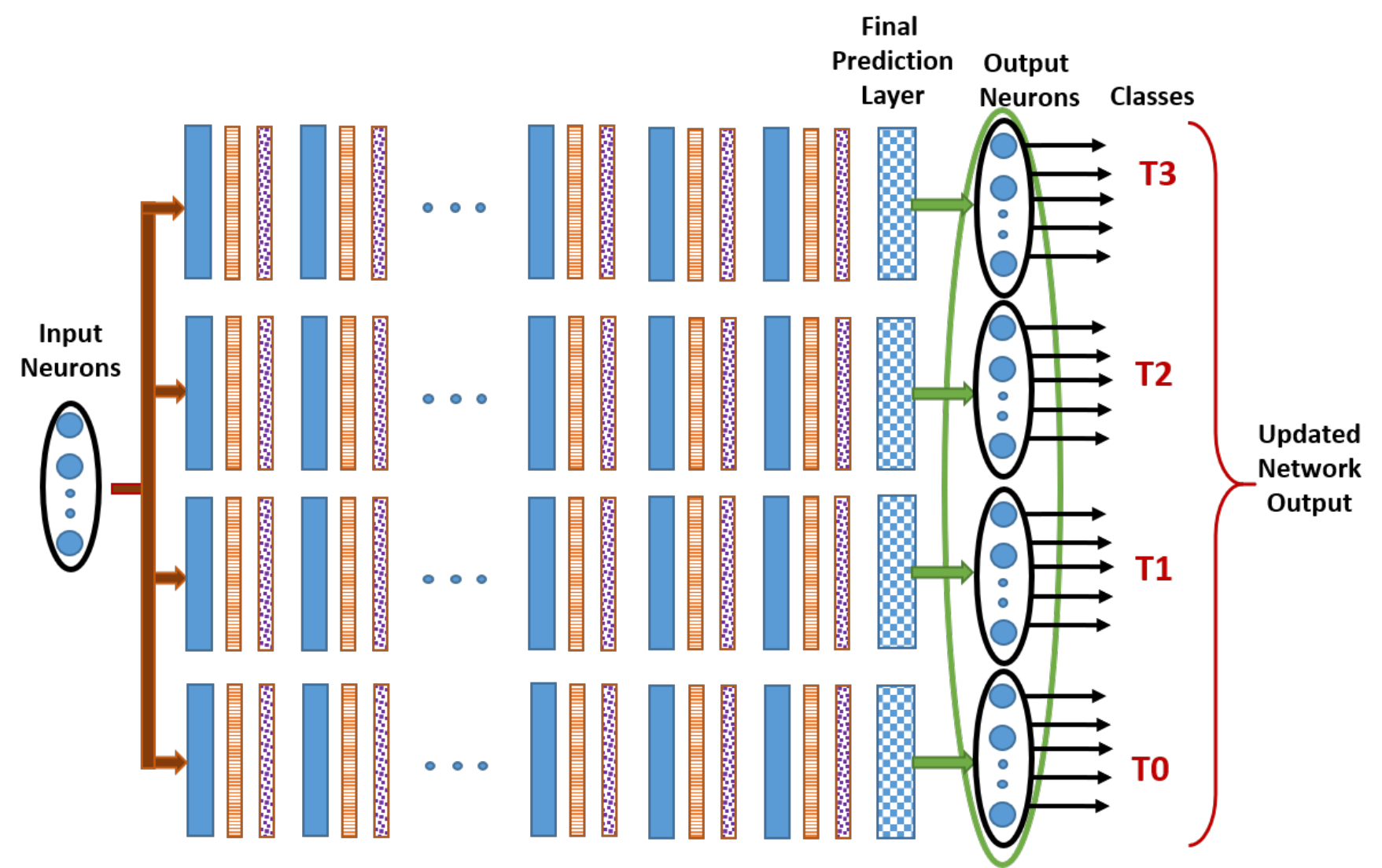}}
\caption{(a) Updated network architecture for training methodology 1. `\%' Sharing is the portion of trainable parameters which are frozen and shared between the base and the new network. This quantity is decided from the \textit{`Accuracy vs Sharing'} curve shown in the inset.  (b) An incrementally trained network, without network sharing, used as baseline for comparison.}
\label{fig38}
\end{figure*}

\subsection{Training Methodology 1}
\label{3.4}
To mitigate the accuracy loss due to sharing, we reduced network sharing and allowed more freedom for retraining the branch networks. By gradually reducing sharing we observed improvement in the inference accuracy for both branch networks and the updated network.\par 
From Figure \ref{fig38a}, we can observe that when we share $\sim$60\% of the learning parameters in the convolutional layers (and corresponding batch normalization and ReLU layers), we can achieve accuracy within $\sim$1\% of baseline. The baseline is an incrementally trained network, without network sharing (Figure \ref{fig38b}). The accuracy results for this network configuration is listed in Table~\ref{Accuracy results for Training Methodology 1}. Note that $\sim$73\%  classification accuracy (row 1, column 4 in Table \ref{Accuracy results for Training Methodology 1}) can be achieved for CIFAR-100 using ResNet50, which is the upper-bound for incremental learning on this network, if the training is done with all training samples applied together. But for incremental learning, all training samples are not available together, hence it is not possible to get that high accuracy even without any network sharing. If we share more than 60\% of the network parameters, classification accuracy degrades drastically. Based on this observation we developed the incremental training methodology for maximum benefits.

\begin{table}[H]
  \caption{Accuracy results for Training Methodology 1}
  \label{Accuracy results for Training Methodology 1}
  \centering
\begin{tabular}{|c|c|c|c|} 
\hline
\multirow{3}{*}{Classes} & \multirow{3}{*}{Network}                                                                                                                                                                        & \multicolumn{2}{c|}{Incremental Learning Accuracy (\%)}                                                                                               \\ \cline{3-4} 
                         &                                                                                                                                                                                                 & \begin{tabular}[c]{@{}l@{}}With partial \\ network sharing\end{tabular} & \begin{tabular}[c]{@{}l@{}}Without partial \\ network sharing\end{tabular} \\ \hline
100 (all classes)                & \multirow{5}{*}{\begin{tabular}[l]{@{}l@{}}ResNet50 ~\cite{he2016deep}:\\ 43 Convolution, \\ 40 Batch Normalization, \\ 40 ReLU, 1 average pooling, \\ 1 Output Prediction layer\end{tabular}} & \textendash                                                                      & 73.95                                                                      \\ \cline{1-1} \cline{3-4} 
50 (base)                &                                                                                                                                                                                                 & \textendash                                                                    & 77.02                                                                       \\ \cline{1-1} \cline{3-4} 
20                       &                                                                                                                                                                                                 & 85.65                                                                    & 85.80                                                                       \\ \cline{1-1} \cline{3-4} 
20                       &                                                                                                                                                                                                 & 84.05                                                                    & 88.00                                                                       \\ \cline{1-1} \cline{3-4} 
10                       &                                                                                                                                                                                                 & 93.50                                                                   & 94.10                                                                       \\ \cline{1-1} \cline{3-4} 
100 (updated)           &                                                                                                                                                                                                 & 59.51                                                                  & 61.00                                                                       \\ 
\hline
\end{tabular}
\end{table}

\begin{figure}[H]
  \centering
  \includegraphics[width=0.9\textwidth]{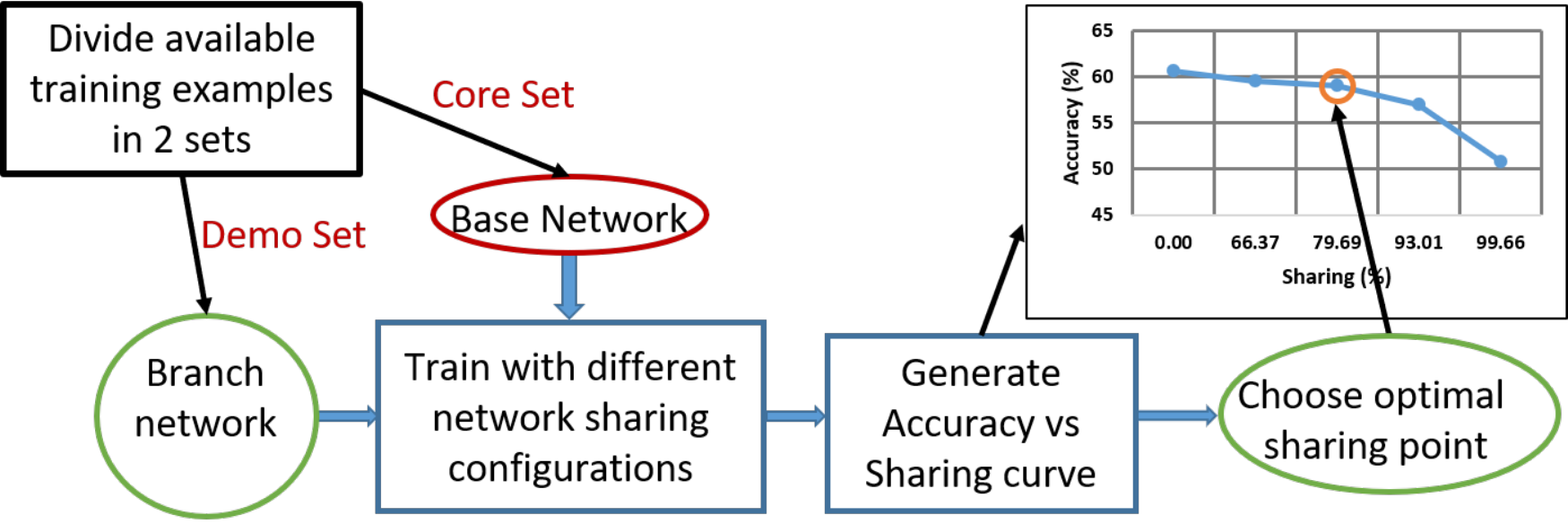}
  \caption{Overview of the DCNN incremental training methodology with partial network sharing.}
  \label{fig39}
\end{figure}

We propose an incremental training methodology with optimal network sharing as depicted in Figure \ref{fig39}. 
The initially available set of classes are divided in to 2 sets. The larger or Core set is used to train a base network. Then the smaller set, which we call a demo set, is used to train a cloned branch network with different sharing configurations. 
From the training results, an {\it Accuracy vs Sharing} curve is generated, from which the optimal sharing configuration for this application and this network architecture is selected. This curve shows how much of the initial layers from the base network can be shared without severe accuracy degradation on the new task. 
An optimal sharing configuration is selected from the curve that meets quality specifications. This optimal configuration is then used for learning any new set of classes.

There is an overhead for determining the optimal sharing configuration from the accuracy-sharing trade-off curve. This curve provides a tuning knob for trading accuracy with energy benefits. However, we do not need to explore the entire search space and we can apply heuristics based on network architecture, number of trainable parameters, dataset complexity, number of training samples etc. to find the optimal sharing configuration within few iterations of retraining the cloned network. Note that training of cloned network is fast and low cost as the shared layers are not backpropagated. In the following paragraph, we will describe the optimal sharing point determination procedure.

To train a base network, we separate initially available classes in two sets: Core set and Demo set. Then we train the base network with the core set and a separate network with the demo set. Accuracy of this separate network will be used as reference for determining the optimal sharing configuration. Next, we create a branch network (that will share some initial layers from the base network) and train it for classes in the demo set. This branch network is a cloned version of the trained base network. The amount of the network sharing can be initially chosen based on the heuristics discussed earlier. For instance, we chose to share 50\% of ResNet50 parameters for CIFAR-100 dataset. Then we train the branch network and compare its performance with the reference accuracy. If the new accuracy is close to the reference, then we increase sharing and train the branch again to compare. On the other hand, if the new accuracy is less than the reference, then we decrease sharing and train again to compare. After few iterations, we finalize the optimal sharing configuration based on the required quality specifications. 
The optimal sharing point is the sharing fraction, beyond which the accuracy degradation with increased sharing is higher than the quality threshold. 
This leads to maximal benefits with minimal quality loss. Finally, we can retrain the base network with both sets (core and demo) together to improve the base network features (in the initial layers), since both sets (core and demo) are available. This base network training and optimal sharing configuration analysis should be done off-chip assuming that there is no energy constraint. Then this base network can be deployed on energy-constrained device (edge) where new classes will be learned. The overhead of optimal sharing configuration selection by generating the accuracy-sharing curve is a onetime cost and it can be neglected since it will be done off-chip.\par

For inference, under the separate task scenario, it will be a two stage network. The multi-stage network will allow selective activation of a task specific branch \cite{panda2017falcon} while other branches will be inactive. For the combined classification scenario, all branches will remain active at the same time.
Note in Figure \ref{fig38a} in the top layers, there are branches for old and new sets of classes. While retraining, and updating the network for a new set of classes, only the branch of top layers corresponding to the new set of classes are retrained. Thus, the top layer filters keep information of their respective tasks (sets of classes) and the network do not suffer from catastrophic forgetting. 

In this work, we do not try to grow a model with classes from datasets of different domains since the base network have learned features from data samples of a single dataset. For instance, if the base network is trained on object recognition dataset CIFAR-10, then it will be able to accommodate new classes from CIFAR-100 dataset as both of the datasets have similar type of basic features (image size, color, background etc.). However, the same base model should not be able to properly accommodate new classes from character recognition dataset (MNIST) because MNIST data has very different type of features compared to CIFAR-10 data. 

\subsection{Training Methodology 2}

\begin{figure}[H]
  \centering
  \includegraphics[width=0.9\textwidth]{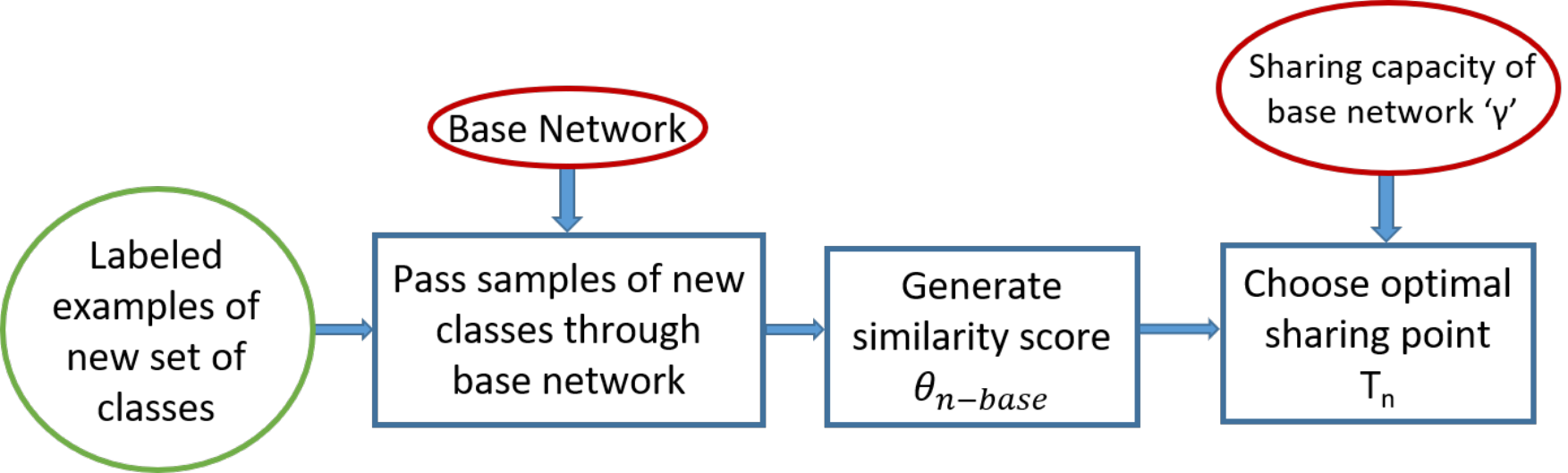}
  \caption{Incremental training methodology for task specific partial network sharing.}
  \label{fig39_2}
\end{figure}

In training methodology 1, all branches of the updated network has equal number of task specific parameters. However, it is not necessary to have this constraint while learning a new task. We extend training methodology 1 to implement task specific sharing configuration as shown in Figure \ref{fig39_2}. In this case, we forward propagate few samples of a new task through the trained base network. From classification results of those samples, we generate a similarity score that approximately quantifies similarity between the classes of base network and the new task. 
To generate the similarity score we have used algorithm \ref{alg:simScore}. We pass random samples of a class belonging to a new task through the trained base network. From the classification results, we count number of repeating classes as similarity points. We repeat this process for all the different classes of the new task several times and take the average number as the similarity score. This is a very simple way to measure similarity, however may not be an ideal one. We employed this method considering its simplicity, so that the overhead of measuring similarity score do not overtake the advantages of task specific sharing. 

\vspace{5mm}
\begin{algorithm}[H]
\SetAlgoNoLine
\KwIn{Trained Base network: Base NN, New task data: TnData.}
\KwOut{Similarity score of new task with respect to learned task: $\theta_{n-base}$}
1. 	Randomly sample 5 training examples for each new class in the new task and forward propagate through the trained base network for classification. 

2. 	Count number of repeating classes as similarity.

3. 	Repeat the steps 1 and 2, 3 times and average the results to get average similarity score.

\caption{Similarity score generation}
\label{alg:simScore}
\end{algorithm}
\vspace{5mm}

Next, we estimate the sharing capacity of the base network from the optimal sharing configuration of the base network and the similarity score of the demo set that was used to generate the `Accuracy-sharing' curve. Then we use equation \ref{eq1} to estimate task specific sharing configuration. The task that have higher similarity with the base network, will be able to share more features from the base network. 
Note that each network architecture has different sharing capacity. Hence, it is necessary to estimate the sharing capacity of the base network using training methodology 1.

\begin{equation}
\label{eq1}
sharing~for~task~`n',~T_n = \gamma\times\theta_{n-base}
\end{equation}
where, $\gamma$ is the sharing capacity of the base network and $\theta_{n-base}$ is the similarity score between the new task and the base network.

\begin{equation}
\label{eq2}
\gamma=\frac{optimal~sharing~of~base~network}{\theta_{base}}
\end{equation}
where, $\theta_{base}$ is the similarity score between the demo set and the core set.

\begin{figure}[H]
  \centering
  \includegraphics[width=0.75\textwidth]{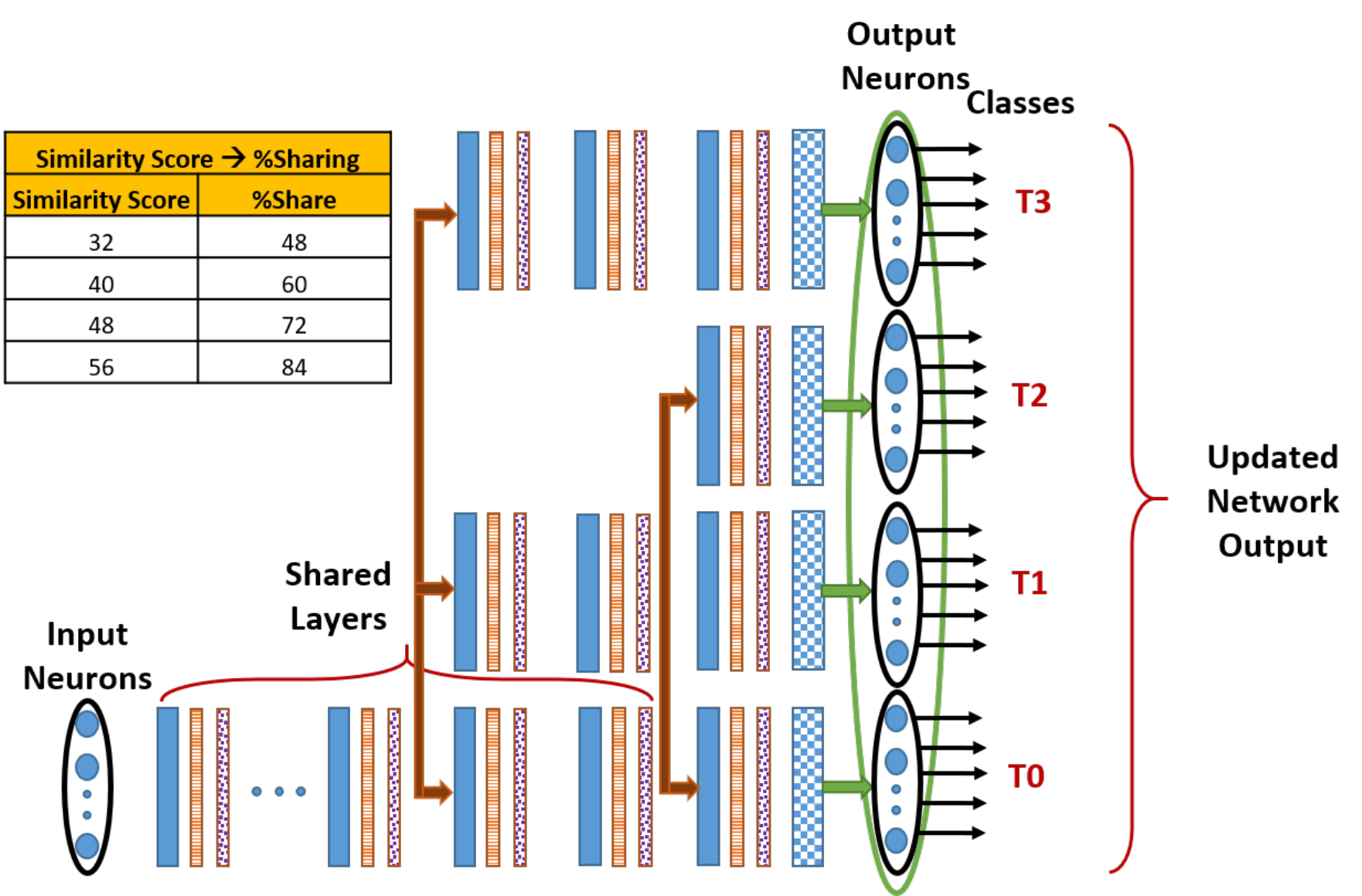}
  \caption{Updated network for task specific partial network sharing using similarity score table. $\gamma$ is equal to 1.5 in the similarity score table.}
  \label{fig39_3}
\end{figure}

Plugging in the network learning capacity and similarity scores in equation \ref{eq1}, we generate a look-up table from which sharing configuration for new task will be determined. The updated network for CIFAR-100 shown in the Figure \ref{fig39_3} is the result of this approach.

\subsection{Training Methodology 3}

\begin{figure}[H]
  \centering
  \includegraphics[width=0.9\textwidth]{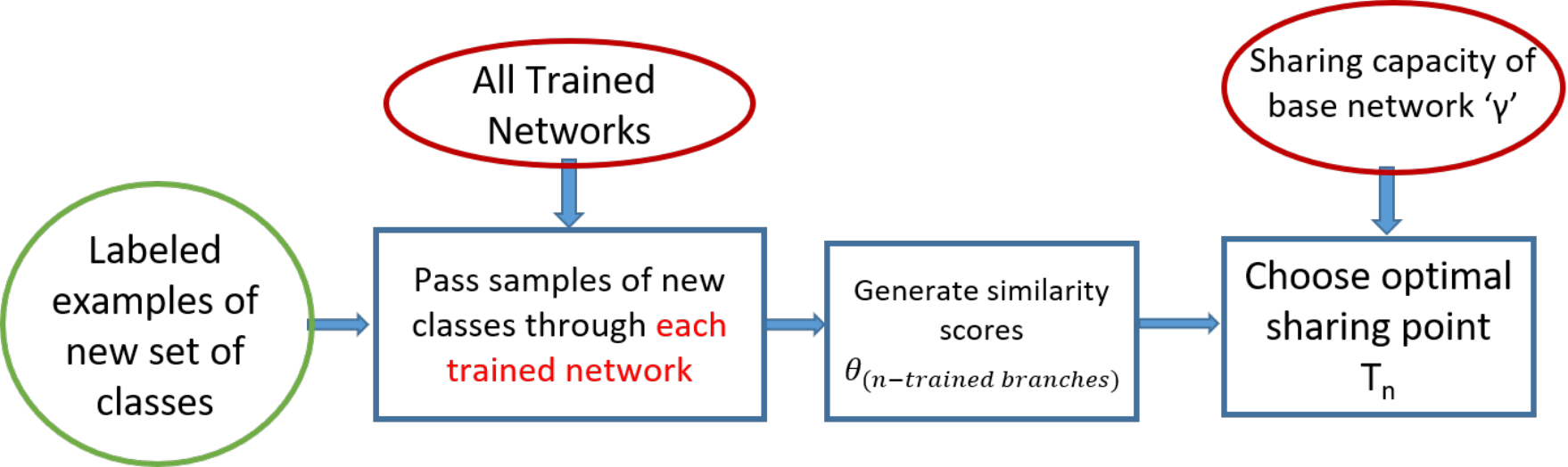}
  \caption{Incremental training methodology for fine grain optimization.}
  \label{fig39_4}
\end{figure}

We can employ a more fine grain optimization by sharing not only from the base network, but also from the trained branches. The training methodology is depicted in Figure \ref{fig39_4}. In this method, we have to generate similarity scores of the new task with all previously learned tasks using algorithm \ref{alg:simScore}. Then, the new task branch will share from the most similar branch based on sharing percentage obtained from \ref{eq3} .

\begin{equation}
\label{eq3}
sharing~for~task~`n',~T_n = \gamma\times~max\theta_{n-trained~networks}
\end{equation}
where, $\gamma$ is the sharing capacity of the base network and $max\theta_{n-trained~networks}$ is the maximum similarity score between new task and trained network branches.

\begin{figure*}[h]
\centering
\subfloat[]{\label{fig39_5a}
  \centering
  \includegraphics[width=.5\textwidth]{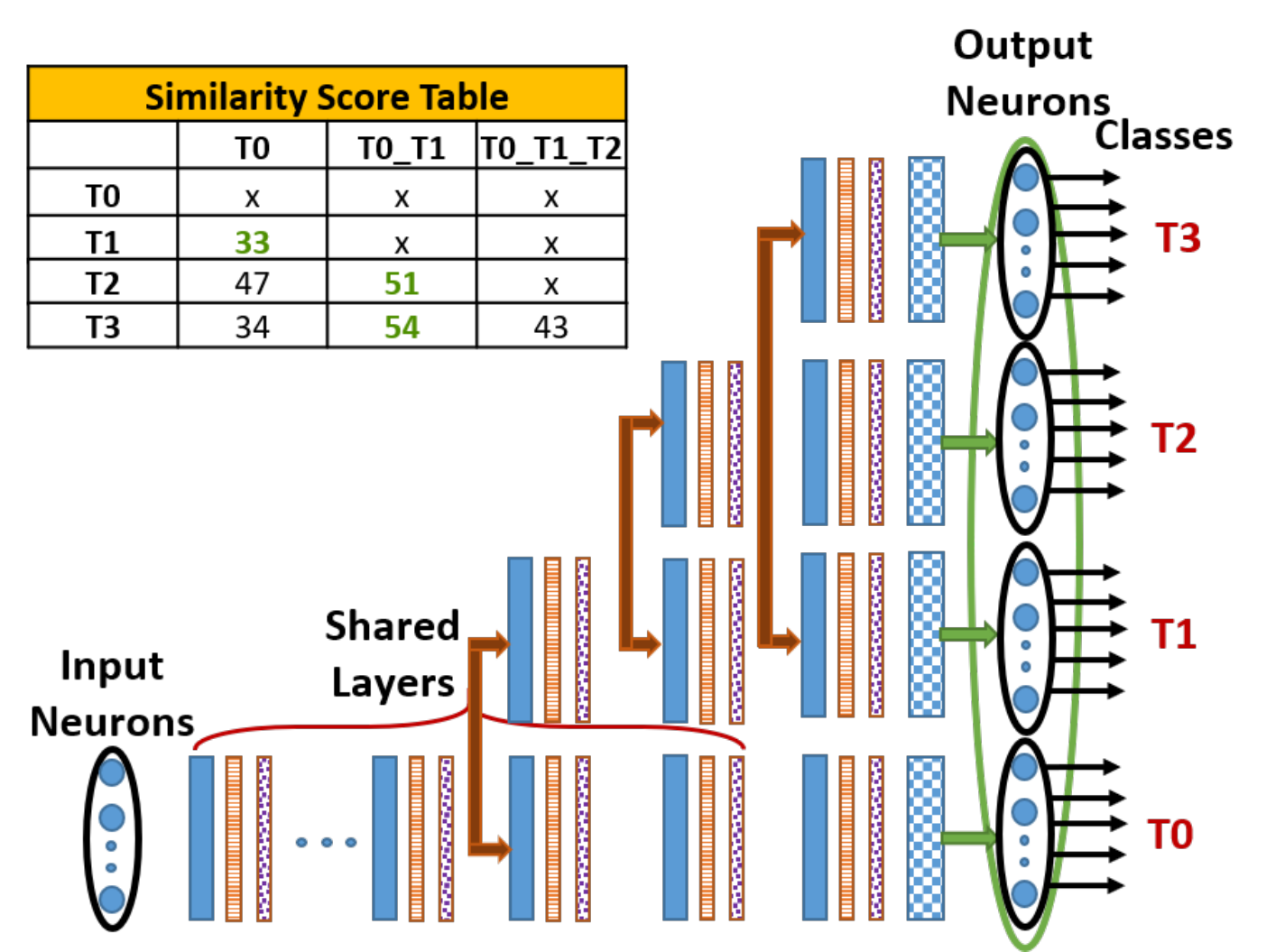}}
\subfloat[]{\label{fig39_5b}
  \centering
  \includegraphics[width=.5\textwidth]{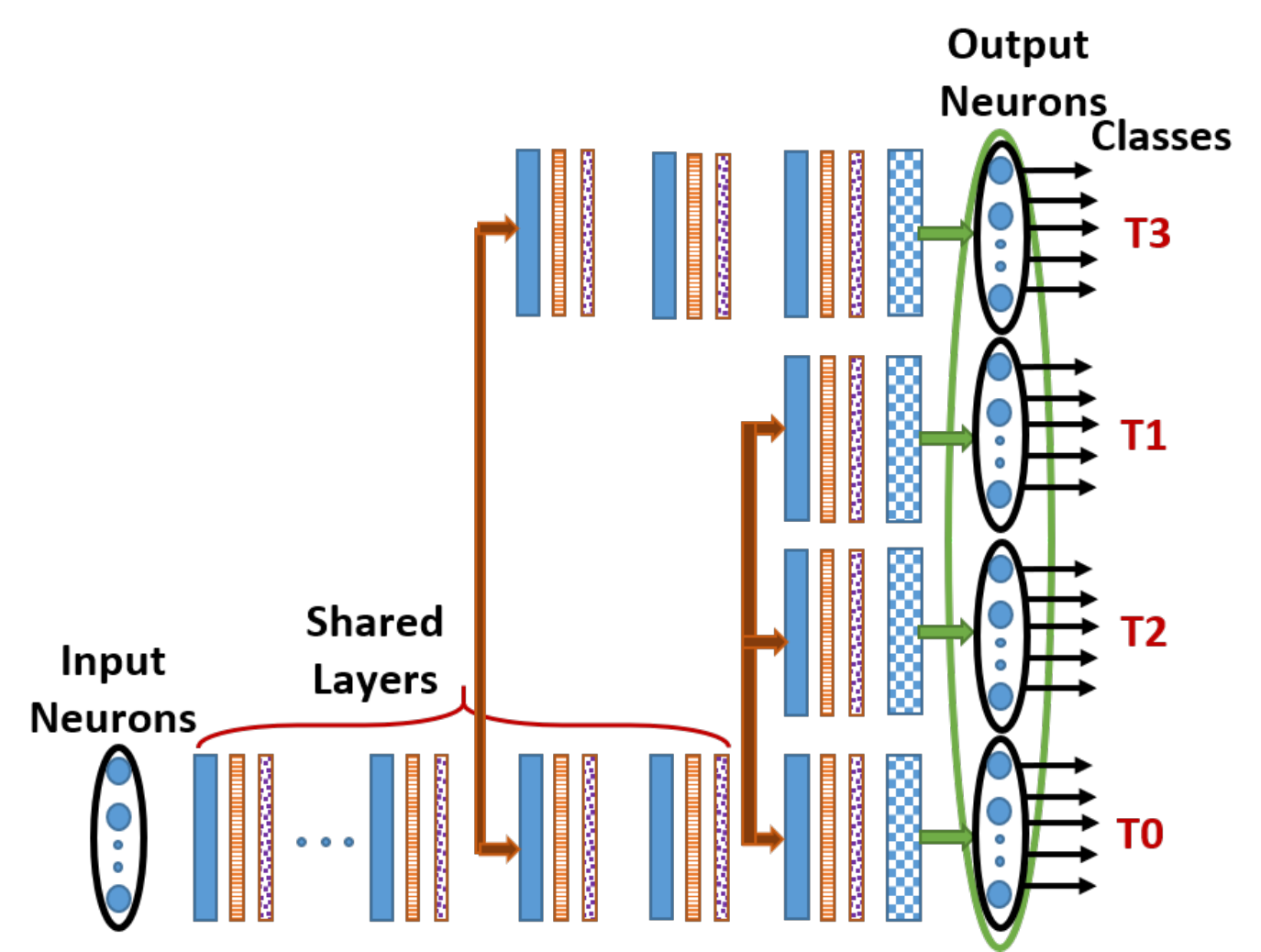}}
\caption{Updated network trained with training methodology 3 for task order (a) T0-T1-T2-T3 and (b) T0-T2-T1-T3.}
\label{39_5}
\end{figure*}

Let us see an example (Figure \ref{fig39_5a}). Assume that the base network is trained with task 0. When data for task 1 becomes available, it has the only option to share network with task 0. Then, data for task 2 arrives and it has two options, share network with task 0 or task 1. From the similarity score table, we can observe that task 2 has higher similarity with task 1 than task 0. So, branch network for task 2 shares network with task 1. For task 3, there are 3 options and it shares with task 1 since it has higher similarity score with task 1.
Note that, task order plays an important role in this approach. For instance, in this specific example, if task 2 is available before task 1 or task 3, we will get a different updated network (Figure \ref{fig39_5b}).  

\subsection{Comparison of Different Training Methodologies}

\begin{table*}[h]
    \caption{Comparison of Different Training Methodologies}
  \label{Training Methodologies}
  \centering
\begin{tabular}{|c|c|c|c|c|c|c|}
\hline
\multirow{2}{*}{Task} & \multicolumn{2}{c|}{Method 1} & \multicolumn{2}{c|}{Method 2} & \multicolumn{2}{c|}{Method 3} \\ \cline{2-7} 
 & Accuracy & Avg. Sharing & Accuracy & Avg. Sharing & Accuracy & Avg. Sharing \\ \hline
T1 & 85.65 & 59.44\% & 86.45 & 46.70\% & 86.45 & 46.70\% \\ \hline
T2 & 84.05 & 59.44\% & 82.30 & 72.20\% & 82.50 & 72.20\% \\ \hline
T3 & 93.50 & 59.44\% & 93.80 & 46.70\% & 93.40 & 84.93\% \\ \hline
T0-T1-T2-T3 & 59.51 & 59.44\% & 60.07 & 55.20\% & 60.58 & 67.94\% \\ \hline
\end{tabular}
\end{table*}

In Table \ref{Training Methodologies}, the task specific and combined classification accuracies are listed with corresponding sharing ratios. For the first training method, we used fixed sharing configuration for all new tasks. In the second training method, we utilized similarity score and found that task 1 and 3 can share less with the base network than task 2. This method reduced the average sharing while increasing the task specific and combined classification accuracy. 
In method 3, task 2 and task 3 were able to share higher amount with task 1 instead of task 0 while maintaining task specific performance. The combined classification is slightly improved while average sharing is also increased.

\section{Evaluation Methodology}
The proposed training approach reduces the number of kernels that would be modified during the training of new classes.
Effectively, this reduces the number of computations in the backward pass, namely layer gradients and weight gradient computations, thereby leading to energy benefits.
The reduction in the number computations is an outcome of the algorithmic optimization \textit{i.e.} reduction in the number of layers during the backpropagation for new classes.
Hence, the energy benefits are not specific to any microarchitectural feature such as dataflow, data reuse etc.
In this work, we use a CMOS digital baseline based on the weight stationary dataflow to analyze the energy consumption for DCNNs.
Weight stationary has been agreed to be an efficient dataflow for executing DNN workloads~\cite{chen2014dadiannao, ankit2019puma}.

The baseline is a many-core architecture where each core maps a partition of the DNN.
Each core is comprised of one or more Matrix Vector Multiplication (MVM) units which perform the MAC operations.
An MVM unit consists of a 32KB memory module with 1024 bit bus width and 32 MACs.
Thus, all the weights (32-bit weight and input) read in a single access from the local memory can be processed in the MACs in one cycle leading to a pipelined execution.
Subsequently, multiple MACs within and across cores operate MVMs in parallel to execute the DNN.
Note that we do not consider the energy expended in off-chip data movement (movement of weight from DRAM to local memory in cores) and inter-core data movement (over network) as these can vary based on the layer configurations, chip size, network design and several optimizations obtained from the software layers~\cite{sze2017efficient}.
We focus only on the compute and storage energy within cores to isolate the benefits derived from the algorithmic features only.
The multiplier and adder unit for MAC was implemented at the Register-Transfer Level (RTL) in Verilog and mapped to IBM 32nm technology using Synopsys Design Compiler, to obtain the energy number.
The memory module in our baseline was modelled using CACTI ~\cite{muralimanohar2009cacti}, in 32nm technology library, to estimate the corresponding energy consumption. 

At the algorithm level, the deep learning toolbox ~\cite{palm2012prediction}, MatConvNet ~\cite{vedaldi2015matconvnet}, and PyTorch ~\cite{paszke2017pytorch}, which are open source neural network simulators in MATLAB, C++, and Python, are used to apply the algorithm modifications and evaluate the classification accuracy of the DCNNs under consideration. The DCNNs were trained, tested and timed using NVIDIA GPUs. In all experiments, previously seen data were not used in subsequent stages of learning, and in each case the algorithm was tested on an independent validation dataset that was not used during training. Details of the benchmarks used in our experiments are listed in Table~\ref{BenchmarksIncre}:

\begin{table*}[h]
  \caption{Benchmarks}
  \label{BenchmarksIncre}
  \centering
\resizebox{\textwidth}{!}{
\begin{tabular}{|c|c|c|}
\hline
Application                       & Dataset          & DCNN Structure                                                                                                                                                     \\ \hline

Object Recog.                  & CIFAR-100 &    ResNet18, ResNet34, ResNet50, ResNet101, DenseNet121 \cite{huang2017densely}, MobileNet \cite{howard2017mobilenets}
\\ \hline
Object Recog.                  & ImageNet &     ResNet34, DenseNet121, MobileNet 
\\ \hline
\end{tabular}}
\end{table*}

\begin{figure*}[h]
\centering
\subfloat[]{\label{fig9a}
  \centering
  \includegraphics[width=.4\textwidth]{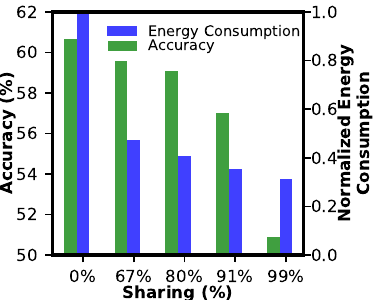}}
\subfloat[]{\label{fig9b}
  \centering
  \includegraphics[width=.4\textwidth]{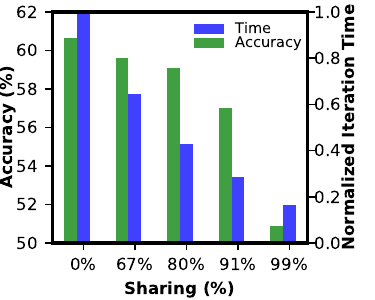}}
\caption{Comparison of (a) energy/accuracy trade-off and (b) training time requirements, between incremental training with and without sharing convolutional layers, is shown for different sharing configurations.}
\label{fig9}
\end{figure*}

\section{Results and Discussions}
In this section, we present results that demonstrate the accuracy obtained, the energy efficiency and reduction in training time, storage requirements and memory access achieved by our proposed design. For these results, we have trained ResNet101 with CIFAR-100. The optimal sharing configuration is 80\%.

\subsection{Energy-Accuracy Trade-off}
DCNNs are trained using the standard back-propagation rule with slight modification to account for the convolutional operators ~\cite{palm2012prediction}. The main power hungry steps of DCNN training (back-propagation) are gradient computation and weight update of the convolutional and fully connected layers ~\cite{sarwar2017gabor}. In our proposed training, we achieve energy efficiency by eliminating a large portion of the gradient computation and weight update operations, with minimal loss of accuracy or output quality. The normalized energy consumption per iteration for incremental training with and without sharing convolutional layers is shown in Figure \ref{fig9a}. The accuracies reported in this work are obtained using test datasets, which are separate from the training datasets. Based on the accuracy requirement of a specific application the optimal sharing point can be chosen from the `Accuracy vs Sharing' curve mentioned in section \ref{3.4}. The optimal sharing configuration for CIFAR100 is 80\% in ResNet101. By sharing 80\% of the base network parameters, we can achieve 2.45x computation energy saving while learning new set of classes. The energy numbers slightly depend on number of classes in the new set of classes to be learned. However, it does not affect much since only the output layer connections vary with the number of new classes, which is insignificant compared to total connections in the network. Note that the energy mentioned in this comparison is computation energy. Memory access energy is discussed in section \ref{MemAccess}.

\subsection{Training Time Reduction}
Since gradient computations and weight updates are not required for the shared convolutional layers, we achieve significant savings in computation time with our proposed approach. Figure \ref{fig9b} shows the normalized training time per iteration for learning a set of new classes. We observe 1.55-6$\times$ reduction in training time per iteration for CIFAR100 in ResNet101 ~\cite{he2016deep} for different sharing configurations. As a byproduct of the proposed scheme, convergence becomes faster due to inheriting features from the base model. Note that the time savings cannot be used to improve accuracy of the networks by providing more epochs to the training. One way to improve accuracy is to retrain the networks with all the training samples (previously seen and unseen), which can be very time consuming and contradictory to the incremental learning principle. 

\begin{figure*}[h]
\centering
\subfloat[]{\label{fig10a}
  \centering
  \includegraphics[width=.4\textwidth]{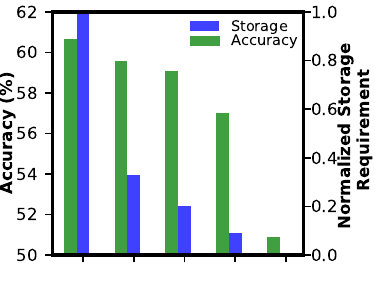}}
\subfloat[]{\label{fig10b}
  \centering
  \includegraphics[width=.4\textwidth]{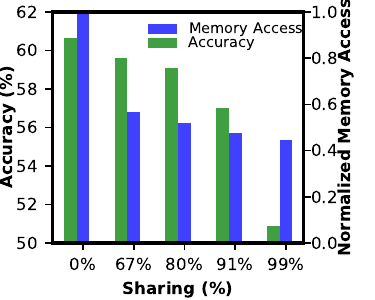}}
\caption{Comparison of (a) storage and (b) memory access requirements, between incremental training with and without sharing convolutional layers, is shown for different sharing configurations.}
\label{fig10}
\end{figure*}

\subsection{Storage Requirement and Memory Access Reduction}
\label{MemAccess}
Figure \ref{fig10a} shows the storage requirement reduction obtained using our proposed scheme for CIFAR100 in ResNet101~\cite{he2016deep}. We achieve 67-99\% reduction in storage requirement since we are sharing initial convolutional layers from the base network for the new branch networks. A large part of the training energy is spent on the memory read/write operations for the synapses. Proposed partial network sharing based training also provides 43-55\% savings in memory access energy during training for CIFAR100 in ResNet101, since we do not need to write (update during backpropagation) the fixed kernel weights during training. Figure \ref{fig10b} shows the memory access requirement reduction obtained using proposed approach.

\begin{table*}[h]
  \caption{Accuracy results for ResNet34 trained on ImageNet}
  \label{table5}
  \centering
\begin{tabular}{|c|c|c|c|c|}
\hline
\multirow{2}{*}{\#Classes} & \multicolumn{2}{c|}{Accuracy(\%) w/o sharing} & \multicolumn{2}{c|}{Accuracy(\%) w/ sharing} \\ \cline{2-5} 
                           & Top 1\%             & Top 5\%            & Top 1\%            & Top 5\%            \\ \hline
1000 (all classes)         & 73.88               & 91.70              & -                  & -                  \\ \hline
500                        & 80.85               & 95.37               & -                  & -                  \\ \hline
300                        & 74.30                & 90.67              & 71.25               & 89.20              \\ \hline
200                 &  75.83                	& 92.61              & 76.60              & 93.12                 \\ \hline
1000 (updated)           & 66.99               & 87.40             & 65.85              & 86.65                   \\ \hline

\end{tabular}
\end{table*}

\subsection{Results on ImageNet}
The ImageNet~\cite{imagenet_cvpr09} (ILSVRC2012) is one of the most challenging benchmarks for object recognition/classification. The data set has a total of \texttildelow1.2 million labeled images from 1000 different categories in the training set. The validation and test set contains 50,000 and 100,000 labeled images, respectively. We implemented ResNet18, ResNet34 ~\cite{he2016deep}, DenseNet121 \cite{huang2017densely} and MobileNet \cite{howard2017mobilenets}, and trained them on ImageNet2012 dataset. We achieved 69.73\%, 73.88\%, 74.23\% and 66.2\% (top 1) classification accuracy for ResNet18, ResNet34, DenseNet121 and MobileNet, respectively, in regular training (all 1000 classes trained together), which are the upper-bounds for the incremental learning on corresponding networks. Then we divided the dataset into 3 sets of 500, 300 and 200 classes for the purpose of incremental learning. The classes for forming the sets were chosen randomly and each set was mutually exclusive. The set of 500 classes were used to train the base network. The other two sets were used for incremental learning. Utilizing our proposed method, we obtained the optimal sharing configuration. 
For ResNet18, we were able to share only \texttildelow1.5\% of the learning parameters from the base network and achieve classification accuracy within $1\pm0.5$\% of the baseline (network w/o sharing) accuracy. On the other hand, using ResNet34, we were able to share up to 33\% of the learning parameters from the base network. The classification accuracy results for ResNet34 with \texttildelow33\% sharing configuration are listed in Table~\ref{table5}. 
This implies that the amount of network sharing largely depends on the network size and architecture. For instance, the DenseNet121 with 121 layers provides \texttildelow57\% sharing, while the MobileNet with only 25 layers provides up to 34\% sharing of the learning parameters (for similar accuracy specifications on ImageNet dataset). Combined classification accuracy achieved on DenseNet121 and MobileNet are \texttildelow64\% and \texttildelow62\%, respectively. The following sub-section analyses the performance and corresponding benefits of proposed methodology on different network architectures.\par

\subsection{Comparison between Different Network Architectures}
\begin{figure*}[h]
\centering
\subfloat[]{\label{fig11a}
  \centering
  \includegraphics[width=.6\textwidth]{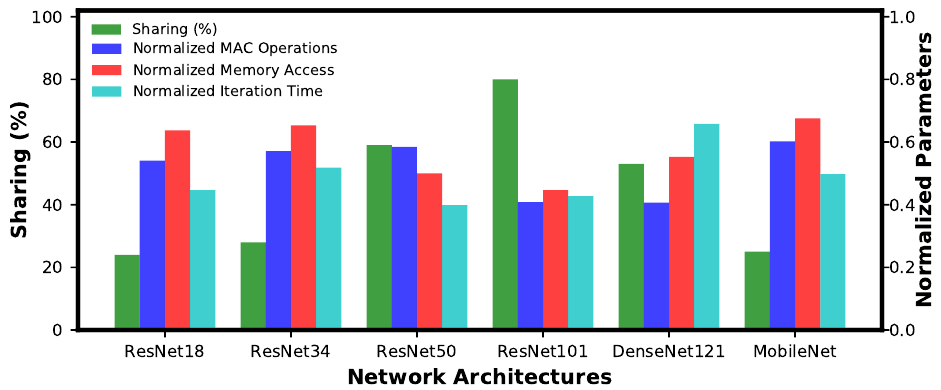}}
\subfloat[]{\label{fig11b}
  \centering
  \includegraphics[width=.4\textwidth]{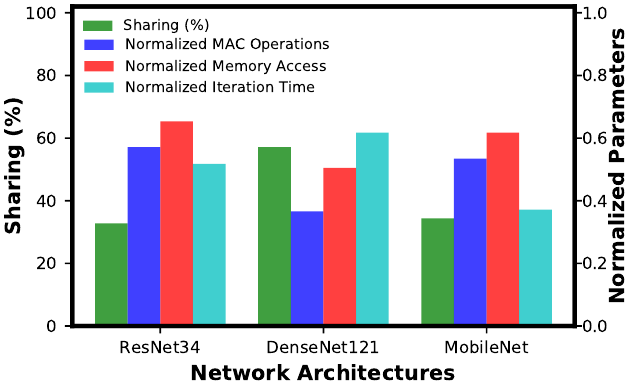}}
\caption{Comparison between different network architectures trained on (a) CIFAR100 and (b) ImageNet, using proposed algorithm. For these experiments, we used $1\pm0.5$\% accuracy degradation as a tolerance value for determining the optimal sharing configuration.}
\label{fig11}
\end{figure*}

We observed that the optimal network sharing configuration depends on network architecture. Therefore, careful consideration is required while selecting the network for using proposed methodology. We experimented with ResNet networks of different depths, DenseNet \cite{huang2017densely} and MobileNet \cite{howard2017mobilenets}. The networks are trained on CIFAR100 and ImageNet with class divisions in Table \ref{Accuracy results for Training Methodology 1} and \ref{table5}. For these experiments, we used $1\pm0.5$\% accuracy degradation as a tolerance value for determining the optimal sharing configuration. Figure \ref{fig11a} shows a comparison between four ResNets (18, 34, 50, 101), Densenet121 and MobileNet network trained with CIFAR-100, while figure \ref{fig11b} shows a comparison between three networks (ResNet34, DenseNet121 and MobileNet) trained for ImageNet dataset. We observed that deeper networks ($>$30 layers) provide more energy benefits for minimal accuracy degradation. However, sharing ratio does not have a linear relation with the energy benefits and training time savings. In Figure \ref{fig11}, we can observe that for similar sharing configurations, different networks achieve different amount of reduction in computational energy, memory access and iteration time. For instance, sharing \texttildelow33\% of the learning parameters in ResNet34, we reduced training time per iteration by 51\% (Figure \ref{fig11b}). On the other hand, similar amount of sharing in MobileNet reduces training time per iteration by 62\% (Figure \ref{fig11b}).\par 
For both CIFAR-100 and ImageNet dataset, MobileNet performs similar to ResNet34 in terms of parameter sharing and energy benefits, while DenseNet provides higher amount of parameter sharing and energy benefits since it has much more depth. 
The trend confirms that there are two prime conditions which need to be satisfied for getting superior performance using the proposed algorithm. Firstly, the network has to be deep enough so that enough layers from base network can be shared. Small networks have most of the weights in the final FC layers which cannot be shared. Larger networks allow more learning parameters to be shared without performance loss. For instance, we could share a lot more of the learning parameters from the base network in ResNet34 compared to ResNet18 for ImageNet. Also for CIFAR-100, percentage sharing is very high in ResNet101 (80\%) compared to ResNet18 (24\%) for an equivalent accuracy degradation. 
Secondly, the base network must contain a good number of features. The combined network performance is best for ImageNet (Table \ref{table5}) compare to CIFAR-100 dataset and much closer to the cumulatively trained network (which is the theoretical upper-bound). This is due to the fact that in the case of ImageNet, the base network has learned sufficient features since it was trained with large number of classes and examples. On the other hand, accuracy is worse for CIFAR-100 (Table \ref{Accuracy results for Training Methodology 1}) as its base network learns only 50 classes and has significantly lower number of training samples.

\begin{table*}[h]
  \caption{Qualitative Comparison with Other Methods}
  \label{table6}
  \centering
\centering
\resizebox{\textwidth}{!}{
\begin{tabular}{|c|c|c|c|c|c|}
\hline
Performance Metric     & Fine Tuning & Feature Extraction & Cumulative Training & Learning w/o Forgetting & This Work \\ \hline
New Task Accuracy      & best        & medium             & best           & best                    & good      \\ \hline
Old Task Accuracy      & worst       & good               & best           & good                    & best      \\ \hline
Training Speed         & fast        & fast               & slow           & fast                    & fastest   \\ \hline
Inference Speed        & fast        & fast               & fast           & fast                    & fast      \\ \hline
Previous Data Required & no          & no                 & yes            & no                      & no        \\ \hline
Storage Requirement    & low         & medium             & highest            & low                     & medium    \\ \hline
\end{tabular}}
\end{table*}

\subsection{Comparison with Other Methods}
Table \ref{table6} presents a qualitative comparison between different methods for incremental learning namely; fine tuning, feature extraction, joint training, Learning without Forgetting \cite{li2016learning} and our proposed method. Here, we considered that accuracy for each task (old and new) is measured separately as reported in \cite{li2016learning}.
In fine tuning, the entire network is retrained for a few epochs to learn the new task.
It suffers from catastrophic forgetting and forgets the old task since old task data is not used during retraining.
In feature extraction, a trained feature extractor is used to extract features for the new task and then a separate classifier is trained on the extracted features.
Although it does not forget the old task, new task performance is lower since the feature extractor is not explicitly trained to extract appropriate features for the new task.
Cumulative or joint training achieves the best accuracy on new tasks without forgetting the old tasks. 
However, it requires old task samples to be stored, which leads to a higher memory requirement compared to the other approaches. 
Also cumulative training is slower compared to other approaches since every retraining utilizes all data samples from old and new tasks. 
`Learning without Forgetting' (LwF) fine tunes the network for new tasks while maintaining response of the new task samples on old task specific neurons~\cite{li2016learning} .
It achieves higher performance on the new task compared to other approaches since the entire network is fine tuned for the new task.
LwF training is fast as it fine tunes the network with new samples only compared to training from scratch using all the task samples.
LwF adds the least number of parameters for the new task~~\cite{li2016learning}, thereby enabling fast inference and lower memory requirements.
LwF aims to achieve energy efficient inference while learning new task with tolerable loss in old task accuracy.
However, a key drawback of LwF is the partial forgetting of old task(s) during the fine-tuning process (learning new-task).
For combined classification, LwF suffers significant accuracy drop.
For instance, ResNet32 incrementally trained in two steps each having 50 classes of CIFAR-100 obtains a top 1 accuracy of \texttildelow52.5\% with LwF \cite{rebuffi2016icarl}.
In a similar setup, our proposed training obtains 62.1\% while reducing the computation energy (\texttildelow60\%), memory access (\texttildelow48\%) and iteration time (\texttildelow57\%) in training. 
Even with periodical utilization of old data samples, \cite{rebuffi2016icarl} reaches 62\% (top 1) accuracy.
\cite{rebuffi2016icarl} also shows that with classes learned in more than two installments, LwF may suffer from continuously degrading performance on old tasks.
On the contrary, we focus on achieving energy efficient training (for new tasks) without tolerating any accuracy degradation in old tasks.
To this effect, we consider scenarios where learning the new task needs to be efficient. 
Consequently, we trade minimal accuracy while achieving maximal energy benefit for the new task. 
Our proposed methodology does not alter the task specific parameters (branch network) for old tasks as well as the shared parameters during the learning of new tasks, thereby not degrading performance for old tasks. 
Further, for the new task we only need to fine-tune the small cloned branch, which makes the training faster compared to other methods.

Task-specific inference cost (energy and time) remains similar for the proposed approach compared to the baseline, as we can activate the specific branch only. 
The combined classification (input can belong to any task) cost will grow sub-linearly as we add more branches, since several branches share initial layer computations. 
An alternative approach to learn new task(s) while retaining the previous task-accuracy would require training separate networks with no weight sharing. 
Subsequently, the inference cost will be maximum in the case of combined classification, as all separate networks would have to be evaluated. Hence, higher sharing improves the efficiency for both training and inference.

In a typical scenario, a model is used for a lot of inference tasks once it is (re-)trained. 
Nonetheless, we focus on facilitating energy constraint training. 
With the advent of emerging algorithms and technology, such scenario will become very popular in near future if on-chip training is made energy-efficient. 
For example, cell phones are increasingly employing on-chip facial and finger print recognition.
In addition to that, on-chip learning will alleviate the requirements to send the data to cloud, thereby enhancing security. 
Similarly, drones can be employed to learn new tasks on the fly without storing the data samples due to memory constraints. In such applications, proposed training algorithm can be beneficial.\par

\begin{figure}[h]
  \centering
  \includegraphics[width=0.5\textwidth]{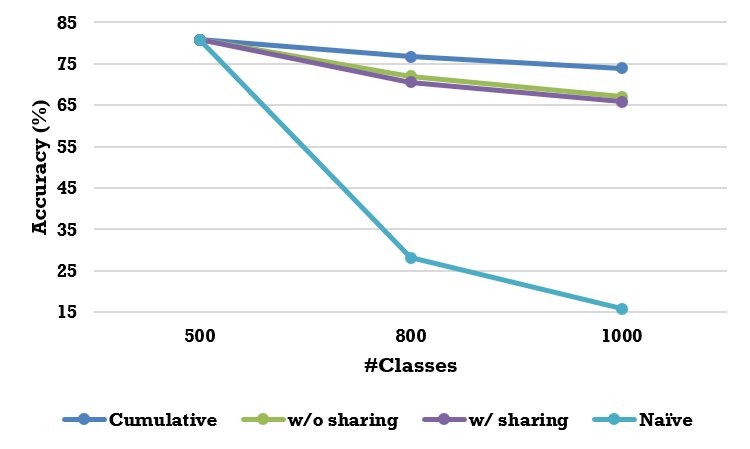}
  \caption{Performance comparison of incremental learning approaches.}
  \label{fig12}
\end{figure}

Next, we will quantitatively compare our approach of incremental learning with two standard approaches in Figure \ref{fig12}. The standard approaches are: \par
\textbf{1. Cumulative:} In this approach the model is retrained with the samples of the new classes and all the previous learned classes (all previous data must be available). This is sort of an upper bound for ideal incremental learning. Since in incremental learning, the old data samples are not available, it will remain an open problem until an approach can match performance of the cumulative approach without using stored data for already learned classes.\par
\textbf{2. Na{\"i}ve:} In this approach, the network is completely retrained with the data samples of new classes only. It suffers from catastrophic forgetting. This approach is also termed as `Fine tuning'.\par

From Figure \ref{fig12}, we can observe that the performance of our proposed method is not very far from the cumulative approach. We also observe that our proposed partial network sharing approach performs almost same as the approach without partial network sharing. While the Na{\"i}ve approach always performs well for the new set of classes only since the network forgets the old classes due to catastrophic forgetting. 

\section{Conclusion}
The performance of DCNNs relies greatly on the availability of a representative set of the training examples. Generally, in practical applications, data acquisition and learning process is time consuming. Also, it is very likely that the data are available in small batches over a period of time. A competent classifier should be able to support an incremental method of accommodating new data without losing ground on old data inference capability. In this paper, we introduce an incremental learning methodology for DCNNs, which employs partial network sharing. This method allows us to accommodate new, previously unseen data without the need of retraining the whole network with previously seen data. It can preserve existing knowledge, and can accommodate new information. Importantly, all new networks start from an existing base network and share learning parameters. The new updated network inherits features from the base network by sharing convolutional layers, leading to improved computational effort and energy consumption during training and thus, speed up the learning process. \par
Our proposed method is simple yet elegant. Most of the other incremental learning approaches focus solely on getting higher accuracy without considering the increase in the complexity of the incremental training process. Most of them retrain the whole network with the new samples. Some of them reuse small percentage of the old training samples during updating the network. In comparison, our approach is an end-to-end learning framework, where we focus on reducing the incremental training complexity while achieving comparable task specific accuracy, and combined classification accuracy close to the upper-bound without using any of the old training samples. Our method does not require any new components other than the branches (regular convolutional and fully-connected layers) for the new classes. The training, updating and inference procedure is the same as the regular supervised learning. This is very useful for re-configurable hardware such as FPGAs where the network capacity can be increased for accommodating new classes without requiring significant effort. In the proposed incremental learning approach, other than showing the learning performance, we directly quantified the effect on energy consumption and memory. We applied the proposed method on different DCNNs trained on real-world recognition applications. Results confirm the scalability of the proposed approach with significant improvements.


%



\section*{Acknowledgment}
The research was funded in part by Center for Brain-inspired Computing (C-BRIC), one of six centers in JUMP, a Semiconductor Research Corporation (SRC) program sponsored by DARPA,  the National Science Foundation, Intel Corporation and Vannevar Bush Faculty Fellowship.





%
\bibliography{main.bbl}

\begin{thebibliography}{10}

\bibitem{aljundi2016expert}
R.~Aljundi, P.~Chakravarty, and T.~Tuytelaars.
\newblock Expert gate: Lifelong learning with a network of experts.
\newblock {\em arXiv preprint arXiv:1611.06194}, 2016.

\bibitem{ankit2019puma}
A.~Ankit, I.~El~Hajj, S.~R. Chalamalasetti, G.~Ndu, M.~Foltin, R.~S. Williams,
  P.~Faraboschi, W.-m. Hwu, J.~P. Strachan, K.~Roy, and D.~Milojicic.
\newblock {PUMA}: A programmable ultra-efficient memristor-based accelerator
  for machine learning inference.
\newblock In {\em International Conference on Architectural Support for
  Programming Languages and Operating Systems (ASPLOS)}, 2019.

\bibitem{chen2014dadiannao}
Y.~Chen, T.~Luo, S.~Liu, S.~Zhang, L.~He, J.~Wang, L.~Li, T.~Chen, Z.~Xu,
  N.~Sun, et~al.
\newblock Dadiannao: A machine-learning supercomputer.
\newblock In {\em Proceedings of the 47th Annual IEEE/ACM International
  Symposium on Microarchitecture}, pages 609--622. IEEE Computer Society, 2014.

\bibitem{chetlur2014cudnn}
S.~Chetlur, C.~Woolley, P.~Vandermersch, J.~Cohen, J.~Tran, B.~Catanzaro, and
  E.~Shelhamer.
\newblock cudnn: Efficient primitives for deep learning.
\newblock {\em arXiv preprint arXiv:1410.0759}, 2014.

\bibitem{imagenet_cvpr09}
J.~Deng, W.~Dong, R.~Socher, L.-J. Li, K.~Li, and L.~Fei-Fei.
\newblock {ImageNet: A Large-Scale Hierarchical Image Database}.
\newblock In {\em CVPR09}, 2009.

\bibitem{fei2006one}
L.~Fei-Fei, R.~Fergus, and P.~Perona.
\newblock One-shot learning of object categories.
\newblock {\em IEEE transactions on pattern analysis and machine intelligence},
  28(4):594--611, 2006.

\bibitem{he2016deep}
K.~He, X.~Zhang, S.~Ren, and J.~Sun.
\newblock Deep residual learning for image recognition.
\newblock In {\em Proceedings of the IEEE conference on computer vision and
  pattern recognition}, pages 770--778, 2016.

\bibitem{howard2017mobilenets}
A.~G. Howard, M.~Zhu, B.~Chen, D.~Kalenichenko, W.~Wang, T.~Weyand,
  M.~Andreetto, and H.~Adam.
\newblock Mobilenets: Efficient convolutional neural networks for mobile vision
  applications.
\newblock {\em arXiv preprint arXiv:1704.04861}, 2017.

\bibitem{huang2017densely}
G.~Huang, Z.~Liu, L.~Van Der~Maaten, and K.~Q. Weinberger.
\newblock Densely connected convolutional networks.
\newblock In {\em Proceedings of the IEEE conference on computer vision and
  pattern recognition}, pages 4700--4708, 2017.

\bibitem{kirkpatrick2017overcoming}
J.~Kirkpatrick, R.~Pascanu, N.~Rabinowitz, J.~Veness, G.~Desjardins, A.~A.
  Rusu, K.~Milan, J.~Quan, T.~Ramalho, A.~Grabska-Barwinska, et~al.
\newblock Overcoming catastrophic forgetting in neural networks.
\newblock {\em Proceedings of the National Academy of Sciences}, page
  201611835, 2017.

\bibitem{krizhevsky2009learning}
A.~Krizhevsky and G.~Hinton.
\newblock Learning multiple layers of features from tiny images.
\newblock Technical report, Citeseer, 2009.

\bibitem{lampert2009learning}
C.~H. Lampert, H.~Nickisch, and S.~Harmeling.
\newblock Learning to detect unseen object classes by between-class attribute
  transfer.
\newblock In {\em Computer Vision and Pattern Recognition, 2009. CVPR 2009.
  IEEE Conference on}, pages 951--958. IEEE, 2009.

\bibitem{li2016learning}
Z.~Li and D.~Hoiem.
\newblock Learning without forgetting.
\newblock In {\em European Conference on Computer Vision}, pages 614--629.
  Springer, 2016.

\bibitem{medera2009incremental}
D.~Medera and S.~Babinec.
\newblock Incremental learning of convolutional neural networks.
\newblock In {\em IJCCI}, pages 547--550, 2009.

\bibitem{mermillod2013stability}
M.~Mermillod, A.~Bugaiska, and P.~Bonin.
\newblock The stability-plasticity dilemma: Investigating the continuum from
  catastrophic forgetting to age-limited learning effects.
\newblock {\em Frontiers in psychology}, 4, 2013.

\bibitem{muralimanohar2009cacti}
N.~Muralimanohar, R.~Balasubramonian, and N.~P. Jouppi.
\newblock Cacti 6.0: A tool to model large caches.
\newblock {\em HP Laboratories}, pages 22--31, 2009.

\bibitem{palm2012prediction}
R.~B. Palm.
\newblock Prediction as a candidate for learning deep hierarchical models of
  data.
\newblock {\em Technical University of Denmark}, 5, 2012.

\bibitem{pan2010survey}
S.~J. Pan, Q.~Yang, et~al.
\newblock A survey on transfer learning.
\newblock {\em IEEE Transactions on knowledge and data engineering},
  22(10):1345--1359, 2010.

\bibitem{panda2018asp}
P.~Panda, J.~M. Allred, S.~Ramanathan, and K.~Roy.
\newblock Asp: Learning to forget with adaptive synaptic plasticity in spiking
  neural networks.
\newblock {\em IEEE Journal on Emerging and Selected Topics in Circuits and
  Systems}, 8(1):51--64, 2018.

\bibitem{panda2017falcon}
P.~Panda, A.~Ankit, P.~Wijesinghe, and K.~Roy.
\newblock Falcon: Feature driven selective classification for energy-efficient
  image recognition.
\newblock {\em IEEE Transactions on Computer-Aided Design of Integrated
  Circuits and Systems}, 36(12), 2017.

\bibitem{paszke2017pytorch}
A.~Paszke, S.~Gross, and S.~Chintala.
\newblock Pytorch, 2017.

\bibitem{patricia2014learning}
N.~Patricia and B.~Caputo.
\newblock Learning to learn, from transfer learning to domain adaptation: A
  unifying perspective.
\newblock In {\em Proceedings of the IEEE Conference on Computer Vision and
  Pattern Recognition}, pages 1442--1449, 2014.

\bibitem{pentina2015curriculum}
A.~Pentina, V.~Sharmanska, and C.~H. Lampert.
\newblock Curriculum learning of multiple tasks.
\newblock In {\em Proceedings of the IEEE Conference on Computer Vision and
  Pattern Recognition}, pages 5492--5500, 2015.

\bibitem{polikar2001learn++}
R.~Polikar, L.~Upda, S.~S. Upda, and V.~Honavar.
\newblock Learn++: An incremental learning algorithm for supervised neural
  networks.
\newblock {\em IEEE transactions on systems, man, and cybernetics, part C
  (applications and reviews)}, 31(4):497--508, 2001.

\bibitem{rebuffi2016icarl}
S.-A. Rebuffi, A.~Kolesnikov, and C.~H. Lampert.
\newblock icarl: Incremental classifier and representation learning.
\newblock {\em arXiv preprint arXiv:1611.07725}, 2016.

\bibitem{1}
C.~Rosenberg.
\newblock {Improving Photo Search: A Step Across the Semantic Gap}.
\newblock
  \url{http://googleresearch.blogspot.com/2013/06/improving-photo-search-step-across.html},
  2013.
\newblock [Online; accessed 26-October-2017].

\bibitem{roy2018tree}
D.~Roy, P.~Panda, and K.~Roy.
\newblock Tree-cnn: A deep convolutional neural network for lifelong learning.
\newblock {\em arXiv preprint arXiv:1802.05800}, 2018.

\bibitem{royer2015classifier}
A.~Royer and C.~H. Lampert.
\newblock Classifier adaptation at prediction time.
\newblock In {\em Proceedings of the IEEE Conference on Computer Vision and
  Pattern Recognition}, pages 1401--1409, 2015.

\bibitem{rusu2016progressive}
A.~A. Rusu, N.~C. Rabinowitz, G.~Desjardins, H.~Soyer, J.~Kirkpatrick,
  K.~Kavukcuoglu, R.~Pascanu, and R.~Hadsell.
\newblock Progressive neural networks.
\newblock {\em arXiv preprint arXiv:1606.04671}, 2016.

\bibitem{sarwar2017gabor}
S.~S. Sarwar, P.~Panda, and K.~Roy.
\newblock Gabor filter assisted energy efficient fast learning convolutional
  neural networks.
\newblock In {\em 2017 IEEE/ACM International Symposium on Low Power
  Electronics and Design (ISLPED)}, pages 1--6, July 2017.

\bibitem{sharif2014cnn}
A.~Sharif~Razavian, H.~Azizpour, J.~Sullivan, and S.~Carlsson.
\newblock Cnn features off-the-shelf: an astounding baseline for recognition.
\newblock In {\em Proceedings of the IEEE conference on computer vision and
  pattern recognition workshops}, pages 806--813, 2014.

\bibitem{shmelkov2017incremental}
K.~Shmelkov, C.~Schmid, and K.~Alahari.
\newblock Incremental learning of object detectors without catastrophic
  forgetting.
\newblock {\em arXiv preprint arXiv:1708.06977}, 2017.

\bibitem{sze2017efficient}
V.~Sze, Y.-H. Chen, T.-J. Yang, and J.~S. Emer.
\newblock Efficient processing of deep neural networks: A tutorial and survey.
\newblock {\em Proceedings of the IEEE}, 105(12):2295--2329, 2017.

\bibitem{vedaldi2015matconvnet}
A.~Vedaldi and K.~Lenc.
\newblock Matconvnet: Convolutional neural networks for matlab.
\newblock In {\em Proceedings of the 23rd ACM international conference on
  Multimedia}, pages 689--692. ACM, 2015.

\bibitem{xiao2014error}
T.~Xiao, J.~Zhang, K.~Yang, Y.~Peng, and Z.~Zhang.
\newblock Error-driven incremental learning in deep convolutional neural
  network for large-scale image classification.
\newblock In {\em Proceedings of the 22nd ACM international conference on
  Multimedia}, pages 177--186. ACM, 2014.

\end{thebibliography}



%







\end{document}